\documentclass[journal]{IEEEtran}
\usepackage[colorlinks,urlcolor=blue,linkcolor=blue,citecolor=blue]{hyperref}
\usepackage{amsmath,amsfonts}
\usepackage{algorithmic}
\usepackage{algorithm}
\usepackage{array}
\usepackage{subfigure}
\usepackage{textcomp}
\usepackage{stfloats}
\usepackage{url}
\usepackage{verbatim}
\usepackage{graphicx}
\usepackage{cite}
\usepackage{booktabs}
\usepackage{algorithm}
\usepackage{algorithmic}
\usepackage{booktabs}
\usepackage{multirow}
\usepackage{amsthm,amsmath,amssymb}
\usepackage{mathrsfs}
\usepackage{xcolor}
\usepackage{newfloat}
\usepackage{listings}
\newfloat{listing}{tb}{lst}{}
\floatname{listing}{Listing}

\usepackage{enumitem}
\newtheorem{theorem}{Theorem}[section]

\theoremstyle{definition}

\usepackage{tikz}

\begin{document}

\title{Frequency-Guided Diffusion Model with Perturbation Training for Skeleton-Based Video Anomaly Detection}

\author{Xiaofeng Tan, Xin Geng,~\IEEEmembership{Senior Member, IEEE}, Liang Wang, \IEEEmembership{Fellow, IEEE}, and Hongsong Wang
        % <-this % stops a space
\thanks{This research was supported by the Jiangsu Science Foundation (BK20230833, BG2024036, BK20243012), the National Science Foundation of China (62302093, 52441503, 62125602, U24A20324, 92464301, 62322607, 62236010, 62276261), the Fundamental Research Funds for the Central Universities (2242025K30024), the Open Research Fund of the State Key Laboratory of Multimodal Artificial Intelligence Systems under Grant E5SP060116, and the Big Data Computing Center of Southeast University. (Corresponding author: Hongsong~Wang)}
\thanks{X. Tan, X. Geng and H. Wang are with the School of Computer Science and Engineering, Southeast University, Nanjing 211189, China, and also with the Key Laboratory of New Generation Artificial Intelligence Technology and Its Interdisciplinary Applications (Southeast University), Ministry of Education, China (email: xiaofengtan@seu.edu.cn; xgeng@seu.edu.cn; hongsongwang@seu.edu.cn). }
\thanks{L. Wang is with New Laboratory of Pattern Recognition (NLPR), State Key Laboratory of Multimodal Artificial Intelligence Systems (MAIS), Institute of Automation, Chinese Academy of Sciences (CASIA), and also with School of Artificial Intelligence, University of Chinese Academy of Sciences (email: wangliang@nlpr.ia.ac.cn).}
}% <-this % stops a space

\markboth{IEEE Transactions on Image Processing,~Vol.~00, No.~0, March~2025}%
{Tan \MakeLowercase{\textit{et al.}}: Frequency-Guided Diffusion for Skeleton-Based Video Anomaly Detection}

\maketitle

\begin{abstract}
Video anomaly detection (VAD) is a vital yet complex open-set task in computer vision, commonly tackled through reconstruction-based methods. However, these methods struggle with two key limitations: (1) insufficient robustness in open-set scenarios, where unseen normal motions are frequently misclassified as anomalies, and (2) an overemphasis on, but restricted capacity for, local motion reconstruction, which are inherently difficult to capture accurately due to their diversity. To overcome these challenges, we introduce a novel frequency-guided diffusion model with perturbation training. First, we enhance robustness by training a generator to produce perturbed samples, which are similar to normal samples and target the weakness of the reconstruction model. This training paradigm expands the reconstruction domain of the model, improving its generalization to unseen normal motions. Second, to address the overemphasis on motion details, we employ the 2D Discrete Cosine Transform (DCT) to separate high-frequency (local) and low-frequency (global) motion components. By guiding the diffusion model with observed high-frequency information, we prioritize the reconstruction of low-frequency components, enabling more accurate and robust anomaly detection. Extensive experiments on five widely used VAD datasets demonstrate that our approach surpasses state-of-the-art methods, underscoring its effectiveness in open-set scenarios and diverse motion contexts. Our project website is \url{https://xiaofeng-tan.github.io/projects/FG-Diff}.
\end{abstract}


\section{Introduction}
\label{sec:intro}

\IEEEPARstart{V}{ideo} anomaly detection (VAD) is dedicated to identifying irregular events within video sequences \cite{cai2021, Luo2021, Cheng2015, Leyva2017, Pham2015}.  Due to the rarity of anomalous events and their inherently ambiguous definitions \cite{Acsintoae2022}, this problem is often considered a challenging task in unsupervised scenarios.  A promising and effective solution \cite{Morais2019, Flaborea2023b, Rai2024} is to train models to capture regular behavioral patterns from normal motions, thereby enabling the identification of deviations as anomalies.  

Based on the data modalities employed, VAD methods \cite{lv2023unbiased,sun2023hierarchical,cao2023new, yang2023video,hirschorn2023normalizing} can be broadly classified into two primary categories: RGB-based \cite{cai2021} and skeleton-based methods \cite{Flaborea2023b}. The former directly processes raw video frames, while the latter utilizes extracted human skeletons, which are less susceptible to noise from illumination changes and background clutter \cite{Mishra2024, WANG2021103225}. Moreover, skeleton-based methods capture low-dimensional, semantically rich features centered on human motion \cite{Morais2019}, making them particularly effective for human-centric VAD.

Generally, existing skeleton-based methods utilize reconstruction \cite{Stergiou2024}, prediction \cite{Rodrigues2020, Slavic2023}, or a combination of both \cite{Flaborea2023b} as auxiliary tasks, to learn regular motion patterns. Among them, reconstruction-based methods are one of the most established methods and have been widely applied in image processing \cite{Chen2024EasyNet}, 3D point cloud analysis \cite{liang2025lookinsidemoreinternal}, and time series modeling \cite{liu2022time}.  In the field of VAD \cite{Wu2024, Liang2023, Shen2024Advancing, Li2022}, Luo et al. \cite{Luo2017Remembering} introduce a reconstruction-based framework for video anomaly detection, enhancing the encoding of appearance and motion regularities in normal events. Astrid et al. \cite{Astrid2021a, Astrid2021b} improve video anomaly detection by training autoencoders with pseudo anomalies generated from normal data to better distinguish normal and anomalous frames. 

\begin{figure}[t]
\centering
\includegraphics[width=0.45\textwidth]{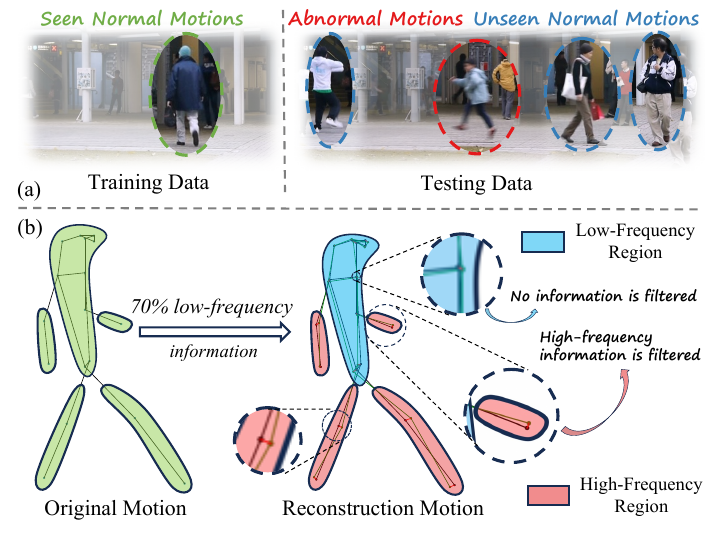}
\caption{The data illustration. (a) The training and testing data, where the training data is composed of seen normal motions and the testing data contains unseen normal and abnormal motions. Although seen and unseen motions represent the same action (e.g., walking), their local details, such as stride length, arm swing amplitude, and joint angles, exhibit significant differences. (b) The frequency analyses of motions. This analysis reveals that a motion retaining only 70\% of its low-frequency information remains largely similar to the original motion in terms of global structure, with minor differences observed in the low-frequency regions. Note that low-frequency and high-frequency regions do not correspond directly to specific joints. Instead, low-frequency regions are defined as areas where joints predominantly contain low-frequency information while also exhibiting a relatively higher proportion of high-frequency details.}
\label{intro_1}
\end{figure}

\noindent\textbf{Motivation.} However, reconstruction-based methods still encounter substantial limitations due to the intrinsic diversity of motion patterns. Specifically, motion patterns are diverse, and even within the same movement category, they may exhibit significant differences in style, amplitude, or speed. Furthermore, we identify two key factors that constrain the performance of reconstruction-based methods, as outlined below.

\textit{A primary limitation of existing methods lies in their inadequate robustness in open-set situations}, where unseen normal samples exhibit subtle differences from those in the training set and often are classified as anomalies. As illustrated in Fig. \ref{intro_1} (a), we observe two phenomena: {(1)} Normal motions in the test set resemble those in the training set, yet they exhibit subtle variations due to individual differences in movement styles and habits. {(2)} Anomalies, in contrast, correspond to irregular actions that deviate from expected behavioral patterns within a specific context, rather than merely exhibiting stylistic variations. However, existing methods primarily capture specific normal motion patterns from limited datasets, limiting their generalization ability for unseen normal motions that exhibit slight stylistic variations. This training paradigm significantly hinders the practical applicability of such models in real-world open-set scenarios.

\begin{figure}[t]
\centering
    \includegraphics[width=0.48\textwidth]{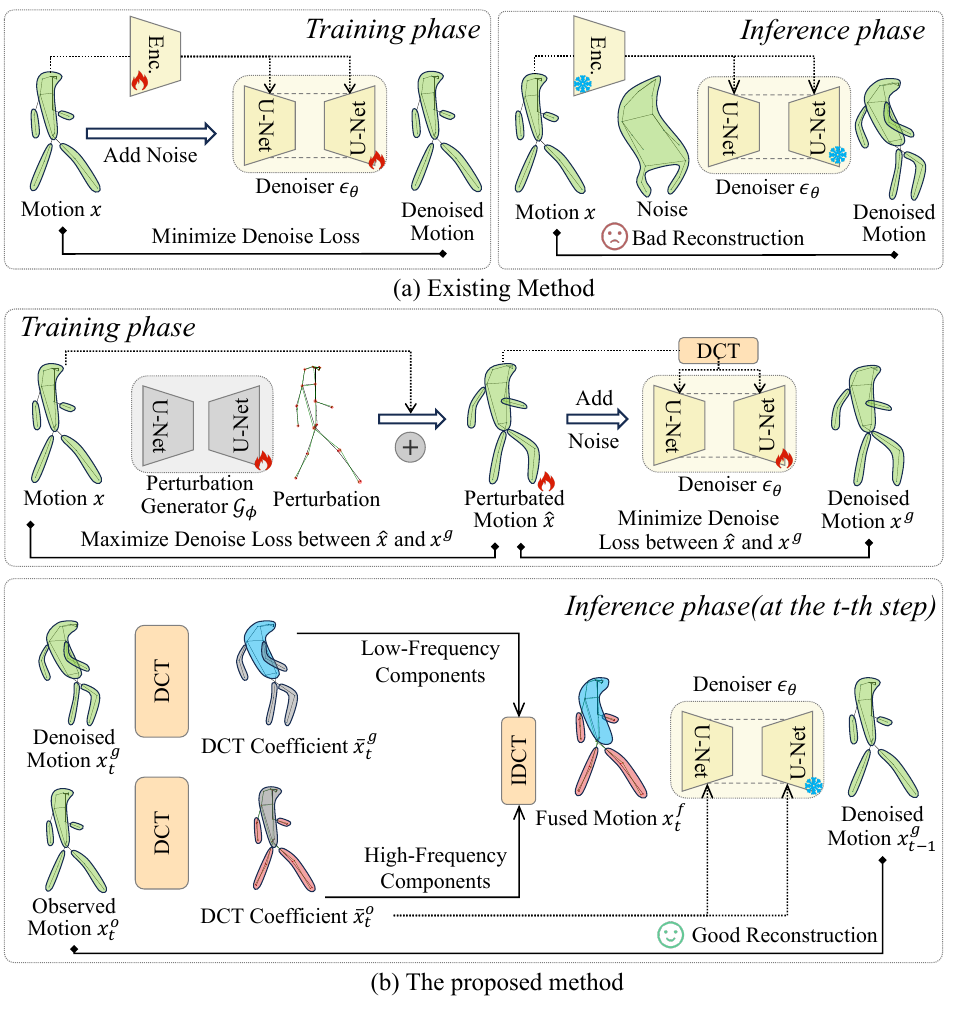}
    \vspace{-3mm} 
        \caption{Comparison between our proposed method and existing methods. During the training phase, we employ adversarial training for the perturbation generator and denoiser to enhance model robustness. Specifically, the perturbation generator attacks the observed motion, producing motions that are challenging to reconstruct yet resemble normal motions. These perturbed motions are then used to train the denoiser, thereby improving its robustness. During the inference phase, we apply DCT to separate observed motion into global and local components, represented as low-frequency and high-frequency information. By leveraging high-frequency information as guidance, our method can accurately reconstruct observed motion compared to existing methods.}
    \label{intro_2}
    \vspace{-15pt}
\end{figure}

\begin{figure}[t]
\centering
    \includegraphics[width=0.48\textwidth]{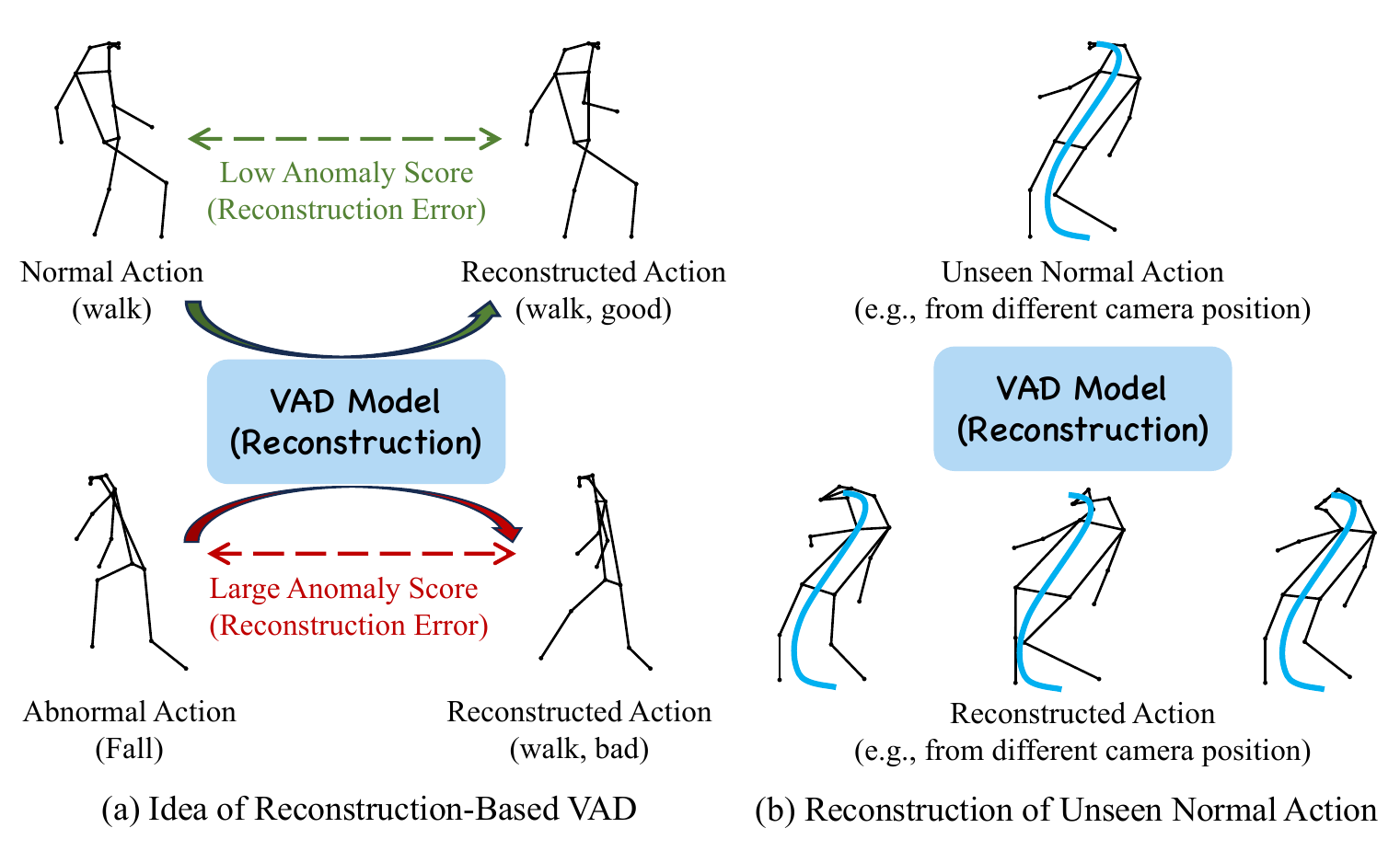}
    \vspace{-3mm} 
        \caption{The illustration of reconstruction-based method. The reconstruction-based method struggles to capture the fine details of unseen actions (e.g., the same action from a different camera view in Fig.(b)) but excels at reconstructing the global action information (as shown by the blue line).}
    \label{fig:rec}
    \vspace{-15pt}
\end{figure}

Secondly, \textit{existing methods typically process both global and local information equally during the inference phase, neglecting their differing contributions.} However, as mentioned above, motions within the same behavioral category share similar global structures yet display significant variations in local details, such as limb movements or speed, due to individual differences (See Fig. \ref{intro_1} (a)).  This means that even though reconstruction-based methods can accurately generate unseen normal motions, their errors in reconstructing local details remain significant due to the inherent diversity of generative models (See Fig. \ref{fig:sde}), particularly in open-set scenarios. In this case, the reconstruction error fails to serve as a reliable indicator for anomaly detection (See Fig. \ref{fig:rec}). A more effective approach would prioritize global information, as it primarily determines the motion category and is critical for anomaly detection.  However, existing methods fail to differentiate the relative importance of these features, limiting their effectiveness in open-set situations.

\noindent\textbf{Contributions.} To address the aforementioned challenges, we propose a \textbf{F}requency-\textbf{G}uided \textbf{Diff}usion (\textbf{FG-Diff}) model with perturbation training, as shown in Fig. \ref{intro_2}. {Firstly}, to enhance the model's robustness against unseen normal motions, we investigate a training paradigm that incorporates perturbation attacks targeting normal motion patterns. This approach aims to expose and mitigate vulnerabilities in the network. Specifically, we train a perturbation generator to produce bounded perturbations, which are restricted to a limited intensity and aim to maximize the reconstruction error. The maximization of reconstruction error highlights the network’s weaknesses in unseen normal motions, and the limited intensity ensures that the perturbed samples remain closely similar to the seen motions.  By integrating such a novel training paradigm, we enhance the model’s generalization capability, thereby addressing the challenge of limited robustness to unseen normal motions. {Secondly}, for the undifferentiated treatment of motion details and global information, our key insight is that these details and global information can be respectively separated as low-frequency and high-frequency components by Discrete Cosine Transform (DCT), as illustrated in Fig. \ref{intro_1} (b). Building upon this observation, we introduce a novel frequency-guided denoising process.  Since high-frequency components are inherently difficult to reconstruct, our approach highlights them as guidance to enhance the reconstruction of overall motions. Specifically, by incorporating high-frequency information extracted from the observed motion, the model prioritizes the reconstruction of low-frequency components, representing the global structure, while preserving critical details. This strategy enhances the differentiation between motion details and global information, overcoming the limitations of prior methods that treat these aspects indiscriminately, as depicted in Fig. \ref{intro_2}.

In this paper, firstly, we introduce a perturbation-based training paradigm for diffusion models to improve robustness against unseen normal motions in open-set scenarios. Secondly, we introduce a frequency-guided denoising process to separate the global and local motion information into low-frequency and high-frequency components, prioritizing global reconstruction for effective anomaly detection. To verify the effectiveness of our method, we conduct extensive experiments on five widely used VAD datasets, demonstrating that the proposed method outperforms state-of-the-art (SoTA) methods.

\section{Related Work}

\subsection{Reconstruction-Based VAD}
As one of the most popular VAD methods, reconstruction-based methods \cite{Hasan2016, Luo2017, Cao2024Context, Astrid2021b}  typically use generative models to learn to reconstruct the samples representing normal data with low reconstruction error. TSC~\cite{Luo2017} uses temporally coherent sparse coding in a stacked recurrent neural network (sRNN) to maintain temporal consistency. To mitigate overfitting in reconstruction-based methods, several works~\cite{Gong2019, Park2020, Liu2021} integrate memory-augmented modules. Gong et al.~\cite{Gong2019} propose MemAE, a memory-augmented autoencoder that constrains reconstruction to normal patterns. Park et al.~\cite{Park2020} employ a memory-augmented strategy to capture normal pattern diversity while limiting network capacity. Liu et al.~\cite{Liu2021} present a hybrid framework combining optical flow reconstruction and frame prediction. Astrid et al.~\cite{Astrid2021a} introduce a temporal pseudo-anomaly synthesizer to train an autoencoder for distinguishing normal and anomalous frames, while their later work~\cite{Astrid2021b} refines this by reconstructing only normal data using pseudo-anomalies. Mishra et al.~\cite{Mishra2024} apply a latent diffusion-based model to generate pseudo-anomalies via inpainting.

\subsection{Skeleton-Based VAD}
Owing to the well-organized structure, semantic richness, and detailed representation of human actions and motion \cite{weng2024usdrl, 11093878, tan2024sopo}, skeletal data has increasingly captivated researchers in video anomaly detection (VAD) over recent years~\cite{wang2025foundation}. Recent advancements in pose-based video anomaly detection (VAD) include eight notable works. Markovitz et al.~\cite{Markovitz2020} project human action graphs into a latent space, using a Dirichlet process mixture for anomaly detection. Flaborea et al.~\cite{Flaborea2023b} apply a diffusion-based generative model, predicting future poses to identify anomalies. COSKAD~\cite{Flaborea2023a} uses one-class classification to map normal motion patterns into a latent space. Hirschorn et al.~\cite{hirschorn2023normalizing} propose a lightweight model based on normalizing flows to minimize nuisance parameters. Zeng et al.~\cite{Zeng2023} introduce HST-GCNN, a hierarchical spatio-temporal graph convolutional network for individual and interpersonal movement analysis. Huang et al.~\cite{Huang2023Hierarchical} develop a hierarchical graph-based framework with a spatio-temporal transformer for body dynamics and interaction modeling. Stergiou et al.~\cite{Stergiou2024} present a multitask framework with an attention-based encoder-decoder for reconstructing occluded skeleton trajectories. Yu et al.~\cite{yu2023regularity} propose a motion embedder and spatial-temporal transformer for self-supervised pose sequence reconstruction. However, these reconstruction-based and skeleton-based methods have not explicitly considered the effect of model robustness and global and local motion information. Therefore, they lack robustness in open-set scenarios due to their inability to generalize to unseen normal motions with subtle stylistic variations. Additionally, they fail to prioritize global information over local details, leading to unreliable reconstruction errors for anomaly detection.

\subsection{Anomaly detection with Perturbations} 
In the field of anomaly detection \cite{TAN2023108971, Tan2025}, several studies have advanced the use of perturbation techniques to enhance the separability of anomalies. Goodfellow et al.~\cite{Ian2014} introduced input perturbation, revealing neural network vulnerabilities to perturbative examples. Leveraging the assumption that normal samples are more sensitive to perturbations, works~\cite{Liang2018, Hsu2020} apply this approach to enhance anomaly separability. Specifically, Liang et al.~\cite{Liang2018} propose a method using subtle input perturbations to distinguish softmax score distributions between normal and abnormal samples. Hsu et al.~\cite{Hsu2020} introduce a preprocessing method that operates without anomaly-specific tuning. However, these methods apply input perturbations only during the testing phase and are not explicitly designed to enhance model robustness.

\subsection{Discrete Cosine Transform for Computer Vision }

 The DCT \cite{Lam2000}, a core technique in signal processing, remains widely utilized across various domains of modern computer vision, particularly those involving temporal data modeling. Recent research has focused heavily on leveraging the DCT for generative tasks. For instance, in human motion modeling  \cite{chen2023humanmac, mao2019learning,yu2025DivDIff,Jeong_2024_CVPR}, Mao et al. \cite{mao2019learning} introduced a natural approach using the DCT to encode temporal information in feed-forward networks for motion prediction. Building upon this, Chen et al. \cite{chen2023humanmac} later leveraged frequency fusion within the DCT space for action prediction. Similarly, in the broader field of video generation, DCTdiff \cite{ning2024dctdiff} operates in the DCT domain to accelerate the generation process. In robotic manipulation, Zhong et al. \cite{zhong2025freqpolicy} utilized the DCT to enhance the precision and smoothness of robot operation.

These studies primarily treat the DCT space as a representation domain to yield smoother, more stable features and faster inference speeds in generative pipelines. In contrast to the prevalent focus on generation, our work focuses on the perspective of reconstruction-based anomaly detection, a representative discriminative task.  We observe that different frequency components contribute disparately to anomaly discrimination. Therefore, our method goes beyond simply adopting the frequency space as a general feature representation. Instead, we explicitly integrate both the high-frequency (body parts) and low-frequency (global movement trends) information derived from the input motion to significantly boost the human abnormal action detection.

\subsection{Diffusion-based VAD}

Early VAD methods were typically grounded in classical generative models, primarily utilizing Variational Autoencoders (VAE) and Generative Adversarial Networks (GANs) as their architectural backbone. However, these architectures present inherent limitations when addressing the challenge of modeling the highly diverse distributions of both normal and anomalous behaviors \cite{Flaborea2023b, liu2025survey}, which is crucial for robust VAD. Specifically, GANs~\cite{Liu2018} are susceptible to the problem of mode collapse \cite{Thanh-Tungcollapse}. This failure limits their ability to accurately represent the complex, multimodal nature of real-world normal motion data. VAEs face a similar constraint~\cite{nguyen2019anomaly}: they can effectively represent only a fixed number of modalities. This fixed capacity limits their ability to capture the  open-set characteristic of anomalous events in reality.

Diffusion Models have recently become a highly promising methodology for these issues \cite{liu2025anomaly}, recognized for their strong capacity to learn complex normal distributions and generate high-quality reconstructions \cite{liu2024Generalized}. Early unsupervised methods, like those by \cite{tur2023exploring}, rely on the premise that anomalies yield high reconstruction errors, a strategy later refined by \cite{tur2023unsupervised} using compact motion representations as conditional information. Recent architectural innovations have significantly enhanced precision: VADiffusion \cite{liu2024vadiffusion} uses a dual-branch structure combining motion vector reconstruction and I-frame prediction, while FDAE \cite{zhu2024flowguided} integrates a flow-guided diffusion autoencoder with sample refinement. Further progress includes methods that enhance the semantic understanding of normal patterns \cite{yan2023feature, cheng2024denoising} and incorporate video depth and optical flow \cite{liang2024movideo}. Considering the recent focus on privacy preservation in surveillance video \cite{liu2025privacy, liuCRCL}, diffusion models are increasingly being applied to skeleton-based methods for VAD. Noteworthy examples include MoCoDAD \cite{Flaborea2023b}, DCMD~\cite{dualcondition25} and GiCiSAD \cite{karami2025graph}. Our method aims to integrate frequency information derived from human skeletons to further boost the performance of skeleton-based anomaly detection.

\section{Preliminaries}
\begin{figure*}[t]
\centering
    \includegraphics[width=\textwidth]{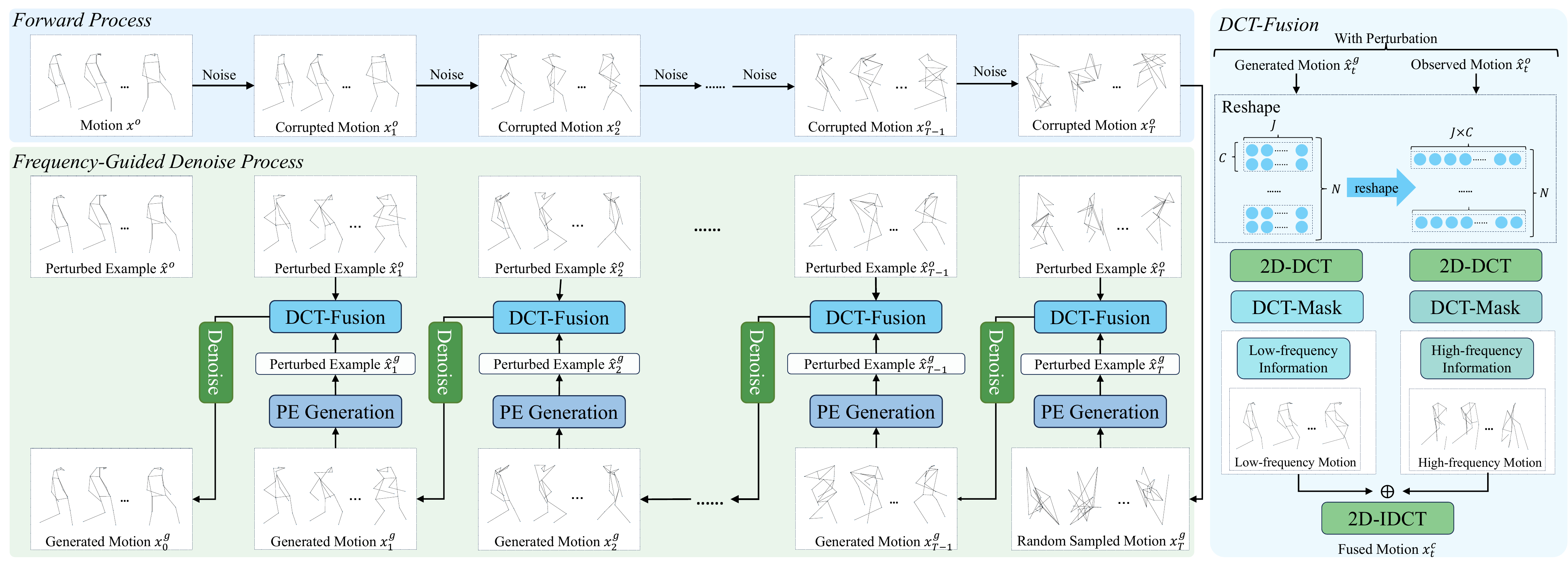}
    \vspace{-8pt}
    \caption{The framework of the proposed method. The model is trained utilizing generated perturbation examples. The training phase includes two processes: minimizing the mean square error to train the noise predictor $\varepsilon_\theta$ and maximizing this error to train the perturbation generator $\mathcal{G}_\phi$. During the testing phase, the high-frequency information of observed motions and the low-frequency information of generated motions are fused for effective anomaly detection.}
    \vspace{-8pt}
    \label{framework}
\end{figure*}

\subsubsection{Diffusion Model for VAD} As a generative model, the diffusion model is trained on normal motions to learn their distribution. During inference, the trained model~\cite{Flaborea2023b} reconstructs motions and assesses anomalies via reconstruction errors. The forward process at timestep \(t\) with variance scheduler \(\alpha_t\) is:
\begin{equation}\label{e1}
\mathbf{x}_t = \sqrt{\Bar{\alpha}_t} \mathbf{x} + \sqrt{1-\Bar{\alpha}_t} \epsilon,
\end{equation}
where $\mathbf{x}_t$ denotes the $N$-frames noised motion sequence $\mathbf{x}^{1:N}$, and $\epsilon$ is a noise from  $\mathcal{N}(\mathbf{0}, \mathbb{I})$. For brevity, the $N$-frames noised motion sequence $\mathbf{x}^{1:N}_t$ at $t$-th step is written as $\mathbf{x}_t$.

During training, the denoiser predicts noise to learn the normal motion distribution:
\begin{equation} \label{eq:dm}
    \mathcal{L}_\mathrm{DM}(\mathbf{x}, \theta) = \mathbb{E}_{\mathbf{x},t} \big[\Vert \epsilon - \epsilon_\theta (\mathbf{x}_t, t, c_\mathbf{x}) \Vert_2^2\big].
\end{equation}
where \(c_\mathbf{x} = \mathrm{Enc}(\mathbf{x})\) is the conditional code encoding motion features via an encoder \(\mathrm{Enc}(\cdot)\).

During inference, motions are reconstructed by denoising:
\begin{equation}
\label{e3}
    \mathbf{x}_{t-1} = \frac{1}{\sqrt{\alpha_t}} (\mathbf{x}_t - \frac{1-\alpha}{\sqrt{1-\Bar{\alpha}}}\epsilon_\theta (x_t, t, c_\mathbf{x})) + (1-\alpha) \epsilon.
\end{equation}

Finally, the generated motion \(\mathbf{x}^g \approx \mathbf{x}_0\) is obtained, and the anomaly score is computed as the reconstruction error:
\begin{equation}\label{eq:rec}
    \mathcal{S}(\mathbf{x}) = \Vert \mathbf{x} - \mathbf{x}^g \Vert_2^2.
\end{equation}

\subsubsection{Discrete Cosine Transform}
In signal processing, the Discrete Cosine Transform (DCT)~\cite{Lam2000} is a key technique for signal transformation. Here, we briefly introduce the 2D-DCT for motion data analysis below. For a motion sequence \(\mathbf{x} \in \mathbb{R}^{N \times C \times J}\), where \(N\) is the number of frames, \(C\) the number of channels, and \(J\) the number of joints, we reshape it into a matrix \(\bar{\mathbf{x}} \in \mathbb{R}^{N \times (C \cdot J)}\), with \(N\) rows for the temporal dimension and \(C \cdot J\) columns for the spatial dimensions. The 2D-DCT and its inverse are defined as \(\mathbf{y} = \mathcal{DCT}(\bar{\mathbf{x}})\) and \(\bar{\mathbf{x}} = \mathcal{IDCT}(\mathbf{y})\), given by:
\begin{equation}
\small
\begin{aligned}
\mathbf{y}_{u,v} &= \alpha(u) \alpha(v) \sum_{i=1}^{N} \sum_{j=1}^{C \cdot J} \bar{\mathbf{x}}_{i,j} \cos \left[ \frac{\pi (2i - 1) u}{2N} \right] \cos \left[ \frac{\pi (2j - 1) v}{2 \cdot C \cdot J} \right], \\
\bar{\mathbf{x}}_{i,j} &= \sum_{u=1}^{N} \sum_{v=1}^{C \cdot J} \alpha(u) \alpha(v) \mathbf{y}_{u,v} \cos \left[ \frac{\pi (2i - 1) u}{2N} \right] \cos \left[ \frac{\pi (2j - 1) v}{2 \cdot C \cdot J} \right],
\end{aligned}
\end{equation}
where the factors \(\alpha(u)\) and \(\alpha(v)\) are:
\begin{equation}
\alpha(u) = 
\begin{cases} 
\sqrt{\frac{1}{N}}, & \text{if } u = 0, \\
\sqrt{\frac{2}{N}}, & \text{otherwise},
\end{cases}
\alpha(v) = 
\begin{cases} 
\sqrt{\frac{1}{C \cdot J}}, & \text{if } v = 0, \\
\sqrt{\frac{2}{C \cdot J}}, & \text{otherwise}.
\end{cases}
\end{equation}

\section{Methodology}

\subsection{Problem Formulation \& Overview}
\noindent\textbf{Problem Settings.} Skeleton-based video anomaly detection is a task to identify abnormal frames containing irregular poses from a given video. Generally, skeleton-based methods first extract human motions $\mathbf{x}^{1:N}= \{\mathbf{x}^1, \mathbf{x}^2, \ldots, \mathbf{x}^N\}$, represented as pose sequences of a fixed length $N$. To simplify the notation, $\mathbf{x}^{1:N}$ is denoted as $\mathbf{x}$. In this step, most existing work adopts the extracted human motion results from preprocessed skeletal datasets \cite{Lu2013, Luo2017, Acsintoae2022}. Next, an anomaly detector is trained to assign a motion-level anomaly score $\mathcal{S}(x)=\{\mathcal{S}(\mathbf{x}^1), \mathcal{S}(\mathbf{x}^2), ..., \mathcal{S}(\mathbf{x}^N)\}$ to each motion $x$. Finally, the frame-level anomaly scores $\mathcal{S}(\mathcal{V})=\{\mathcal{S}(f^1), \mathcal{S}(f^2), ..., \mathcal{S}(f^v)\}$ are obtained through post-processing according to motion-level anomaly scores $\mathcal{S}(x)$, where $\mathcal{V}=\{f^1, f^2, ..., f^v\}$ is a $v$-frames video and $f^i$ denotes the $i$-th frame. Our work primarily focuses on obtaining motion-level anomaly scores $\mathcal{S}(\mathbf{x})$ for each motion using the proposed frequency-guided diffusion model.

\noindent\textbf{Overview.} In response to the issues mentioned in Sec. \ref{sec:intro},  we propose a frequency-guided diffusion model with perturbation training, as illustrated in Fig.~\ref{framework}. To enhance robustness against unseen normal motions, we introduce a training paradigm where a perturbation generator \(\mathcal{G}_\phi\) produces bounded perturbations on motions \(\mathbf{x}\), maximizing reconstruction error to expose network vulnerabilities while keeping perturbed samples \(\hat{\mathbf{x}}\) similar to seen motions. This perturbation training alternates with a noise predictor \(\epsilon_\theta\), improving the generalization of the model. Additionally, to address the undifferentiated treatment of motion details and global information, our model leverages DCT to separate low-frequency (global) and high-frequency (local) components. During inference, the frequency-guided denoising process fuses low-frequency information from generated motion \(\hat{\mathbf{x}}_t^g\) with high-frequency information from observed motion \(\hat{\mathbf{x}}_t^o\), prioritizing the reconstruction of global structures while preserving critical details.

\subsection{Diffusion Model with Perturbation Training}

\begin{figure}[tbp]
\centering
    \includegraphics[width=0.45\textwidth]{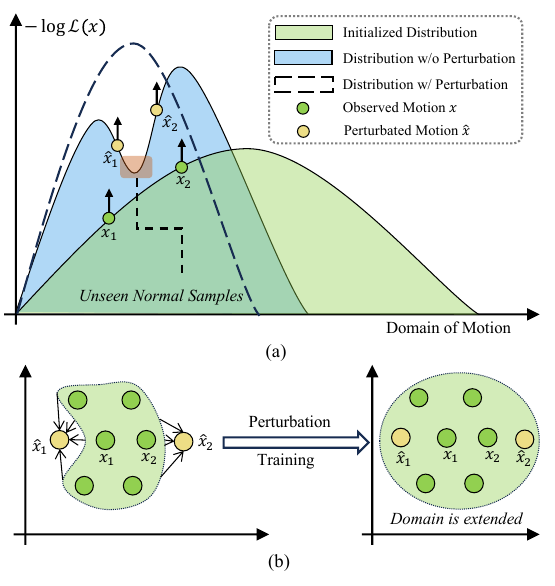}
    \vspace{-5pt}
    \caption{The illustration of perturbation training. In Fig. (a), the green and yellow points denote the original training $x_k$ and perturbed motion $\hat{x}_k$, respectively. The red region represents the distribution of unseen normal samples. Accordingly, Fig. (b) demonstrates that the reconstruction domain is extended by our proposed perturbation training.}
    \label{perturbed}
    \vspace{-10pt}
\end{figure}

\noindent \textbf{Motivation.} Existing reconstruction-based video anomaly detection (VAD) methods primarily focus on reconstructing seen normal motions, yet they exhibit limited robustness when encountering unseen normal motions. As depicted in the blue region of Fig.~\ref{perturbed}(a), these methods are typically trained on a limited set of observed normal motions \(\mathbf{x}^o\), which restricts their ability to generalize due to the absence of unseen normal training motions. Consequently, unseen normal motions, represented in the red region of Fig.~\ref{perturbed}(a), are frequently misclassified as anomalies, underscoring a significant limitation in their generalization capability.

To overcome this challenge, we aim to enhance model robustness by identifying and training on potential unseen normal motions derived from limited observed normal motions \(\mathbf{x}^o\). These potential unseen normal motions, denoted as \(\hat{\mathbf{x}}^o\), should closely resemble observed normal motions \(\mathbf{x}^o\) while inducing a larger reconstruction error to expose the model’s weaknesses. Formally, given a neighborhood parameter \(\lambda\) and network parameters \(\theta\), the ideal potential unseen normal motions should satisfy the following conditions:
\begin{enumerate}[label=(\alph*)]
    \item \textit{Similarity to observed normal motions}: \(\Vert \mathbf{x}^o - \hat{\mathbf{x}}^o \Vert \leq \lambda\);
    \item \textit{Increased reconstruction error}: \(\mathcal{S}(\mathbf{x}^o) - \mathcal{S}(\hat{\mathbf{x}}^o) \leq 0\).
\end{enumerate}

To this end, we propose a perturbation training approach for diffusion-based models. Drawing inspiration from adversarial examples~\cite{Ian2014}, our key insight is to expand the model’s reconstruction domain by generating perturbed examples, enabling it to better handle unseen normal samples, as illustrated in Fig.~\ref{perturbed}(b). Specifically, we introduce a small perturbation \(\delta\) to a given normal motion \(\mathbf{x}^o\), producing a potential unseen normal motion \(\hat{\mathbf{x}}^o\), which enhances the model’s ability to generalize across a broader range of normal motion patterns.

\noindent \textbf{Perturbed Motion Generation.} To generate these perturbed motions, we aim to find a perturbation \(\delta\) that maximizes the loss function while remaining within a constrained neighborhood. This is formulated as:
\begin{equation}\label{eq:1}
    \begin{aligned}
        \delta = \arg \max_{\delta \in \mathcal{N}(\mathbf{x}^o, \lambda_p)} \mathcal{L}(\mathbf{x}^o + \delta, \theta),
    \end{aligned}
\end{equation}
where \(\mathcal{N}(\mathbf{x}^o, \lambda_p)\) denotes the norm constraint with a maximum perturbation intensity \(\lambda_p\), and \(\theta\) represents the model parameters.

Inspired by the fast gradient sign method (FGSM)~\cite{Ian2014}, the perturbation \(\delta\) can be approximately computed as:
\begin{equation}\label{eq:fgsm}
    \begin{aligned}
        \delta = \lambda_p \mathrm{sign}\big(\nabla_{\mathbf{x}^o} \mathcal{L}(\mathbf{x}^o, \theta)\big).
    \end{aligned}
\end{equation}
where $\mathrm{sign}(\cdot)$ denotes the sign function. In this case, the perturbed motion is then constructed as:
\begin{equation} \label{eq:x}
    \begin{aligned}
\hat{\mathbf{x}}^o = \mathbf{x}^o + \delta. 
    \end{aligned}
\end{equation}
By design, \(\hat{\mathbf{x}}^o\) remains similar to \(\mathbf{x}^o\) due to the small perturbation budget \(\lambda_p\), yet it induces a higher loss, making it an effective sample for exposing the model’s vulnerabilities to unseen normal motions. Training on such perturbed motions enables the model to better distinguish between normal and anomalous patterns, thereby improving its robustness.

\noindent \textbf{Perturbation Generator.}
However, directly computing gradients for diffusion models at each iteration, as in Eq.~\eqref{eq:fgsm}, is computationally expensive and memory-intensive. To address this, we introduce a lightweight perturbation generator \(\mathcal{G}_\phi\), parameterized by \(\phi\), to efficiently predict the optimal perturbation \(\delta_\phi\) at a reduced computational cost. The perturbation in this way is generated as:
\begin{equation}\label{eq:4}
    \begin{aligned}
        \delta_\phi &= \lambda_p \mathrm{sign}\big(\mathcal{G}_\phi (\mathbf{x}^o)\big),
    \end{aligned}
\end{equation}
Here, the perturbation generator \(\mathcal{G}_\phi\) is optimized to maximize the model’s loss, thereby exposing vulnerabilities to unseen normal motions, as formulated by:
\begin{equation}\label{eq:5}
    \begin{aligned}
        &\max_{\phi} \mathcal{L}(\mathbf{x}^o +\delta_\phi, \theta),\\
    \end{aligned}
\end{equation}
By training \(\mathcal{G}_\phi\) to produce effective perturbations, the diffusion model learns to handle perturbed motions \(\hat{\mathbf{x}}_\phi\) that mimic potential unseen normal motions, enhancing its robustness in a computationally efficient manner.

\begin{algorithm}[t]
    \caption{Perturbation Training for Diffusion Model}
    \label{AT}
    \textbf{Input}: The observed motions \(\mathbf{x}^o\), the noising steps \(T\), the maximum iterations \(I_{\text{max}}\) \\
    \textbf{Output}: The noise predictor \(\epsilon_\theta\), the perturbation generator \(\mathcal{G}_\phi\)
    \begin{algorithmic}[1]
    \STATE Encode the conditional code: \(c_\mathbf{x} = \mathcal{DCT}_k(\mathbf{x}^o)\)
    \FOR{\(i = 1, 2, 3, \ldots, I_{\text{max}}\)}
    \STATE Sample the timestep \(t\) from \(\mathcal{U}_{[1,T]}\)
    \STATE Sample Gaussian noise \(\epsilon\) from \(\mathcal{N}(\mathbf{0}, \mathbb{I})\)
    \STATE Add noise \(\epsilon\) to \(\mathbf{x}^o\) using variance scheduler \(\alpha_t\): \(\mathbf{x}^o_t = \sqrt{\bar{\alpha}_t} \mathbf{x}^o + \sqrt{1 - \bar{\alpha}_t} \epsilon\)
    \STATE Generate perturbed example using the perturbation generator: \(\hat{\mathbf{x}}^o_t = \mathbf{x}^o_t + \lambda_p \mathrm{sign}(\mathcal{G}_\phi (\mathbf{x}^o_t, t))\)
    \STATE Compute the noise prediction loss: \(\mathcal{L}(\hat{\mathbf{x}}^o, \theta, \phi) = \mathbb{E}_{\hat{\mathbf{x}}^o, t} \big[\Vert \epsilon - \epsilon_\theta (\hat{\mathbf{x}}^o_t, t, c) \Vert_2^2\big]\)
    \STATE Freeze the parameters of \(\mathcal{G}_\phi\) and update \(\epsilon_\theta\) by minimizing \(\mathcal{L}(\hat{\mathbf{x}}^o, \theta, \phi)\)
    \STATE Repeat the process from lines 4 to 7
    \STATE Freeze the parameters of \(\epsilon_\theta\) and update \(\mathcal{G}_\phi\) by maximizing \(\mathcal{L}(\hat{\mathbf{x}}^o, \theta, \phi)\)
    \ENDFOR
    \end{algorithmic}
\end{algorithm}

\begin{theorem} [Effectiveness of perturbation generator]\label{the:1}
Given an observed motion $\mathbf{x}^o$, a perturbation generator \(\mathcal{G}_\phi\) trained by Eq. (\ref{eq:5}), and a neighborhood parameter \(\lambda\), the generated perturbed motion $\hat{\mathbf{x}}^o$ obtained by Eq.  (\ref{eq:x}) and Eq. (\ref{eq:4})  satisfies that:
\begin{enumerate}[label=(\alph*)]
    \item {Similarity to observed normal motions}: \(\Vert \mathbf{x}^o - \hat{\mathbf{x}}^o \Vert \leq \lambda\);
    \item {Increased reconstruction error}: \(\mathcal{S}(\mathbf{x}^o) - \mathcal{S}(\hat{\mathbf{x}}^o) \leq 0\).
\end{enumerate}
\end{theorem}
The proof is provided in the Appendix \ref{App}. Theorem~\ref{the:1} confirms that \(\mathcal{G}_\phi\) generates perturbed motions that enhance robustness by ensuring similarity to observed motions while increasing reconstruction error.

\noindent \textbf{Perturbation Training for Diffusion Model.} 
During training, the parameters of the diffusion model \(\epsilon_\theta\) are continuously updated, leading to evolving vulnerabilities in its performance. To address this challenge, we propose an adversarial training framework that dynamically optimizes both the diffusion model \(\epsilon_\theta\) and the perturbation generator \(\mathcal{G}_\phi\). Specifically, \(\epsilon_\theta\) is trained to minimize the loss function \(\mathcal{L}(\theta, \mathbf{x}^o)\), while \(\mathcal{G}_\phi\) is optimized to maximize \(\mathcal{L}(\theta, \hat{\mathbf{x}}^o)\), where \(\mathbf{x}^o\) and \(\hat{\mathbf{x}}^o\) are constrained to remain similar. This adversarial optimization is formally expressed as:
\begin{equation}
    \min_{\theta} \max_{\phi} \mathcal{L}(\mathbf{x}^o + \lambda_p \mathrm{sign}(\mathcal{G}_\phi (\mathbf{x}^o)), \theta, \phi),
\end{equation}
where the loss function \(\mathcal{L}(\theta, \hat{\mathbf{x}}^o)\) extends Eq.~\eqref{eq:dm} by incorporating the perturbation generator \(\mathcal{G}_\phi\), defined as:
\begin{equation}
    \begin{aligned}
        \mathcal{L}(\mathbf{x}^o, \theta, \phi) = \mathbb{E}_{\mathbf{x}^o, t} \big[\Vert \epsilon - \epsilon_\theta (\hat{\mathbf{x}}^o_t, t, c_{\mathbf{x}^o}) \Vert_2^2\big],
    \end{aligned}
\end{equation}
with \(\mathbf{x}^o_t\) defined by Eq.~\eqref{e1}. Here, \(c_{\mathbf{x}^o}\) denotes the conditional code, derived by selecting the top \(k\) largest DCT coefficients \(\mathcal{DCT}_k(\mathbf{x}^o)\), and the perturbed motion \(\hat{\mathbf{x}}^o_t\) is given by:
\begin{equation}
    \hat{\mathbf{x}}^o_t = \mathbf{x}^o_t + \lambda_p \mathrm{sign}(\mathcal{G}_\phi (\mathbf{x}^o_t, t)).
\end{equation}

In summary, the proposed framework adversarially optimizes the perturbation generator and the noise predictor during training, with the detailed procedure outlined in Algorithm~\ref{AT}.

\begin{figure}[tbp]
\centering
    \includegraphics[width=0.45\textwidth]{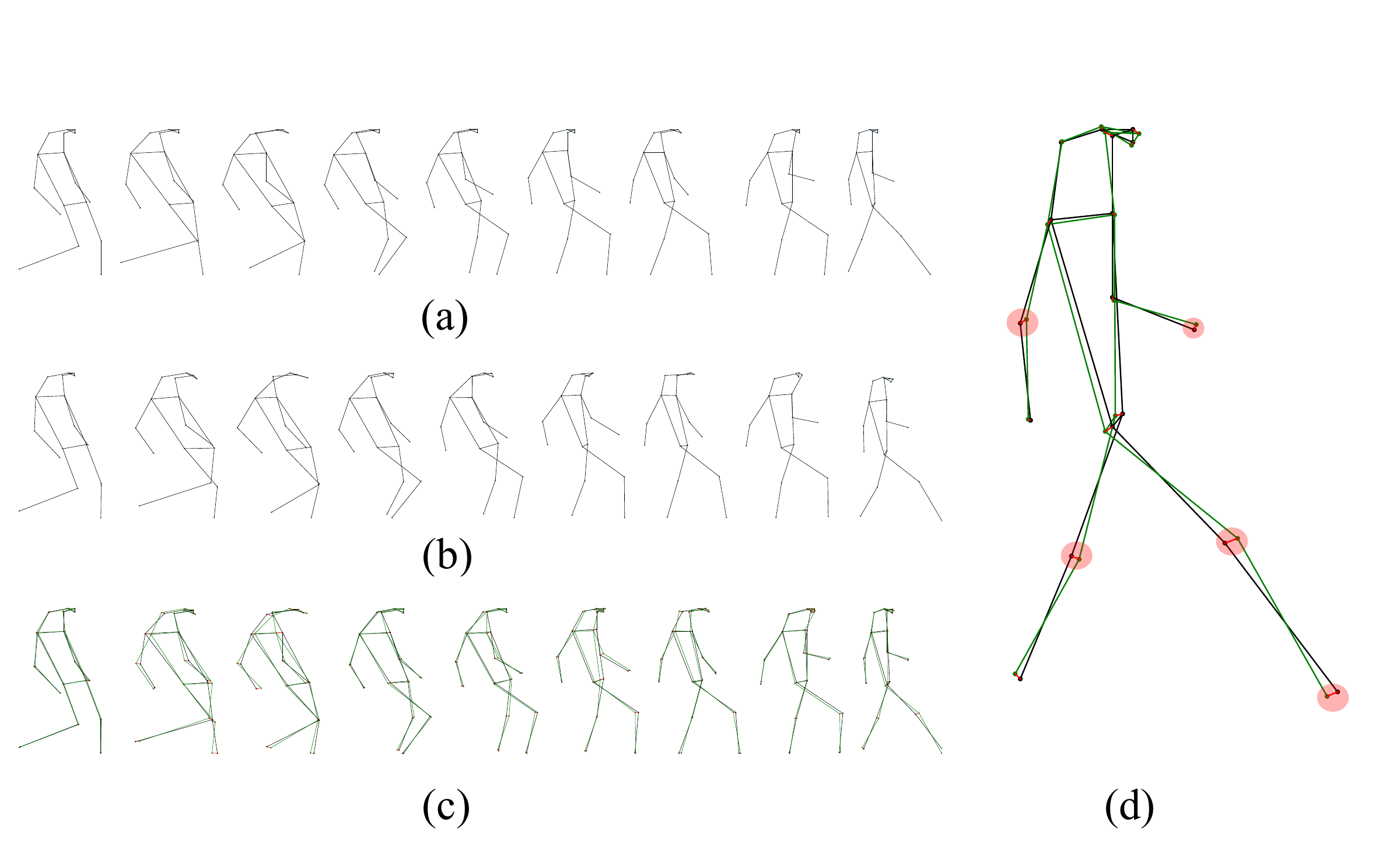}
    \caption{The visualization of human motions processed by 2D-DCT. (a) original motions; (b) motions with low-frequency information only; (c) the comparison between (a) and (b); (d) the skeletal example. Note that the red lines in (d) denote the discarded high-frequency information, and the red circles represent the high-frequency joints w.r.t. temporal and spatial dimension.}
    \label{dct-e}
    \vspace{-6pt}
\end{figure}

\subsection{Frequency-Guided Motion Denoise Process}
\noindent\textbf{Frequency Information in Motion.} In signal processing, high-frequency information refers to rapid variations or fine details, while low-frequency information represents slower changes or broad features. Similarly, the \textit{low-frequency information} in human motion provides basic outlines of behavior, e.g., the center of gravity, the gesture pose, and action categories. In contrast, \textit{high-frequency information} captures details of the motion. Owing to the diversity of personal habits, high-frequency information tends to vary from person to person, such as the stride length and the extent of hand swing while walking. As shown in Fig. \ref{dct-e}(a), (b), and (c), the motions containing only low-dimensional information are almost identical to the original motions,  except for only a few joints. A closer examination of these joints in Fig. \ref{dct-e}(d) reveals that most differences are derived from personal habits, such as the degree of knee bending when walking. In this case, the reconstruction quality, especially that of the joints with high-frequency information, is no longer a reliable indicator for anomaly detection. 

Generative models struggle to accurately reconstruct high-frequency motion details due to their diversity, but this should not deem the generated motions unrealistic. Instead, accurate reconstruction of low-frequency information, combined with rich high-frequency details, indicates that the motion aligns with the expected distribution and is not an anomaly. However, existing methods indiscriminately prioritize all frequency information equally, which compromises detection accuracy.

To this end, we propose a frequency-guided denoising process for anomaly detection, comprising three key steps: (1) frequency information extraction, (2) separation of high-frequency and low-frequency components, and (3) frequency information fusion. 

\noindent \textbf{Frequency Information Extraction.} 
To capture both temporal and spatial characteristics, we employ the 2D-DCT for frequency information extraction. The original motion \(\mathbf{x}\) is first reshaped into a condensed form \(\bar{\mathbf{x}} \in \mathbb{R}^{N \times (C \cdot J)}\), where \(N\) represents the temporal dimension, \(C\) the number of channels, and \(J\) the number of joints. The 2D-DCT and its inverse (IDCT) are applied as follows:
\begin{equation}
    \mathbf{y} = \mathcal{DCT}(\bar{\mathbf{x}}), \quad \bar{\mathbf{x}} = \mathcal{IDCT}(\mathbf{y}),
\end{equation}
where \(\mathbf{y} \in \mathbb{R}^{N \times (C \cdot J)}\) denotes the DCT coefficients obtained from the transformed motion \(\bar{\mathbf{x}}\).

\noindent \textbf{Frequency Information Separation.} 
To separate frequency components, we introduce a DCT-Mask that isolates low-frequency information from DCT coefficients \(\mathbf{y}\). The low-frequency mask \(\mathcal{M}_l \in \{0, 1\}^{N \times (C \cdot J)}\) is defined as:
\begin{equation}
    [\mathcal{M}_l(\mathbf{y})]_{i,j} =
    \begin{cases}
        1, & \text{if } |\mathbf{y}_{i,j}| \geq \tau, \\
        0, & \text{otherwise},
    \end{cases}
\end{equation}
where \(\tau\) is a threshold determined by the top \(\lambda_{\text{dct}}\) percent largest absolute value among all DCT coefficients in \(\mathbf{y}\), ensuring that only the most significant low-frequency components are retained. Similarly, the high-frequency mask \(\mathcal{M}_h \in \{0, 1\}^{N \times (C \cdot J)}\) is defined as \(\mathcal{M}_h(\mathbf{y}) = \mathbf{1} - \mathcal{M}_l(\mathbf{y})\), capturing the complementary high-frequency information.

\begin{algorithm}[tb]
    \caption{Frequency-Guided Motion Denoising Process}
    \label{DCT-If}
    \textbf{Input}: The noise predictor \(\epsilon_\theta\), the perturbation generator \(\mathcal{G}_\phi\), the noising steps \(T\), the perturbation magnitude \(\lambda_p\) \\
    \textbf{Output}: Generated motion \(\mathbf{x}^g_0\)
    \begin{algorithmic}[1]
    \STATE Encode the conditional code: \(c_\mathbf{x} = \mathcal{DCT}_k(\mathbf{x}^o)\)
    \STATE Sample Gaussian noise \(\mathbf{x}_t^g \sim \mathcal{N}(\mathbf{0}, \mathbb{I})\)
    \FOR{\(t = T, T-1, T-2, \dots, 1\)}
    \STATE Sample Gaussian noise \(\varepsilon \sim \mathcal{N}(\mathbf{0}, \mathbb{I})\) if \(t \neq 1\); else set \(\varepsilon = \mathbf{0}\)
    \STATE Add noise \(\varepsilon\) to \(\mathbf{x}^o\) using variance scheduler \(\alpha_t\): \(\mathbf{x}_t^o = \sqrt{\bar{\alpha}_t} \mathbf{x}^o + \sqrt{1 - \bar{\alpha}_t} \varepsilon\)
    \STATE Generate adversarial examples using the perturbation generator: \(\hat{\mathbf{x}}_t^o = \mathbf{x}_t^o + \lambda_p \mathrm{sign}(\mathcal{G}_\phi (\mathbf{x}_t, t))\), \(\hat{\mathbf{x}}_t^g = \mathbf{x}_t^g + \lambda_p \mathrm{sign}(\mathcal{G}_\phi (\mathbf{x}_t^g, t))\)
    \STATE Reshape the observed motion \(\hat{\mathbf{x}}_t^o\) and generated motion \(\hat{\mathbf{x}}_t^g\) into condensed forms \(\hat{\bar{\mathbf{x}}}_t^o\) and \(\hat{\bar{\mathbf{x}}}_t^g\)
    \STATE Transform the condensed motions into the DCT domain: \(\mathbf{y}_{t}^{o} = \mathcal{DCT}(\hat{\bar{\mathbf{x}}}_t^o)\), \(\mathbf{y}_{t}^{g} = \mathcal{DCT}(\hat{\bar{\mathbf{x}}}_t^g)\)
    \STATE Fuse the observed and generated motions using DCT masks: \(\mathbf{y}_{t}^{c} = \mathbf{y}_{t}^{o} \odot \mathcal{M}_h(\mathbf{y}_{t}^{o}) + \mathbf{y}_{t}^{g} \odot \mathcal{M}_l(\mathbf{y}_{t}^{g})\)
    \STATE Transform and reshape the fused motion back to the original space: \(\mathbf{x}_{t}^{c} = \text{reshape}(\mathcal{IDCT}(\mathbf{y}_{t}^{c}))\)
    \STATE Denoise the motion using the formula: 
    \[\small
    \mathbf{x}_{t-1}^g = \frac{1}{\sqrt{\alpha_t}} \left( \mathbf{x}_{t}^{c} - \frac{1 - \alpha_t}{\sqrt{1 - \bar{\alpha}_t}} \epsilon_\theta (\mathbf{x}_{t}^{c}, t, c) \right) + (1 - \alpha_t) \varepsilon
    \]
    \ENDFOR
    \STATE \textbf{return} The generated motion \(\mathbf{x}_0^g\)
    \end{algorithmic}
\end{algorithm}

\noindent\textbf{Pipeline \& Frequency Information Fusion.} The proposed model utilizes frequency information by fixing the high-frequency components and combining them with low-frequency components to generate motions accurately. Given a motion $\Bar{\mathbf{x}}_{t}$ corrupted from observation, and $\Bar{\mathbf{x}}_{t}^d$ generated from the denoising process, the corresponding frequency information is obtained by 2D-DCT:
\begin{equation}
        \mathbf{y}_{t}^{o} = \mathcal{DCT}(\Bar{\mathbf{x}}_{t}^o), \;\;
        \mathbf{y}_{t}^{g} = \mathcal{DCT}(\Bar{\mathbf{x}}_{t}^g).
\end{equation}
Then, the DCT coefficients are masked with the DCT-Mask and combined into a fused coefficient:
\begin{equation}
        \mathbf{y}_{t}^{c} =\mathbf{y}_{t}^{o} \odot \mathcal{M}_h(\mathbf{y}_{t}^{o}) + \mathbf{y}_{t}^{g} \odot \mathcal{M}_l(\mathbf{y}_{t}^{g}).
\end{equation}
Subsequently, the fused coefficients are transformed into the original space by IDCT:
\begin{equation}
        \Bar{\mathbf{x}}_{t}^{c} =  \mathcal{IDCT}(\mathbf{y}_{t}^{c}).
\end{equation}
Finally, the coefficient $\Bar{\mathbf{x}}_{t}^{c}$ is reshaped to the motion $\mathbf{x}_{t}^{c}$. 

During this process, the fused motion is obtained by fusing high-frequency components of observation with low-frequency components of generation. Furthermore, the denoised motions of the $t-1$ step are given as: 
\begin{equation}
\label{e16}
    \mathbf{x}_{t-1}^{g} = \frac{1}{\sqrt{\alpha_t}} (\mathbf{x}_{t}^{c} - \frac{1-\alpha}{\sqrt{1-\Bar{\alpha}}}\epsilon_\theta (\mathbf{x}_{t}^{c}, t, c)) + (1-\alpha_t) \epsilon.
\end{equation}
Finally, the anomaly score can be obtained by Eq. \ref{eq:rec}. We describe the frequency-guided motion denoising process in Algorithm \ref{DCT-If}. Specifically, the frequency-guided motion denoising process first encodes the conditional code of the input motion using the DCT. Then, Gaussian noise is sampled and added to the motion data using the variance scheduler. Subsequently, for each time step,  the motion data is corrupted by the Gaussian noise. Next, the adversarial example is generated using the perturbation generator and both the observed and generated motions are transformed into DCT space. Finally, the observed and generated motion data are combined using a DCT-Mask, and then the fused motion data are converted back to the original space by IDCT.

\section{Experiment}
\begin{table*}[t] \small 
\caption{Comparison of the proposed method against other SoTA methods. The best results across all methods are in bold, the second-best ones are underlined, ``Pred." and ``Rec." denote {predict-based} and {reconstruction-based} methods, respectively.}
\centering
\begin{tabular}{clclccccc}
\toprule
        \textbf{Type} &\textbf{Method} & \textbf{Venue} & \textbf{Modality} & \textbf{Avenue} & \textbf{HR-Avenue} &\textbf{HR-STC} & \textbf{UBnormal} &\textbf{HR-UBnormal} \\ 
\midrule
        \multirow{7}*{Pred.}  & MPED-RNN-Pred. \cite{Morais2019} & \textit{CVPR' 2019} & Skeleton & - & - & 74.5 & - & - \\ 
        ~& Multi-Time. Pred. \cite{Rodrigues2020} & \textit{WACV' 2020} & Skeleton & - & 88.3 & 77.0 & - & - \\ 
        ~& PoseCVAE \cite{Jain2020} & \textit{ICPR' 2021} & Skeleton & - & 87.8 & 75.7 & - & - \\ 
        ~& AMMC \cite{cai2021} & \textit{AAAI' 2021} & RGB & 86.6 & - & - & - & - \\ 
        ~& F$^2$PN \cite{Luo2022} & \textit{T-PAMI' 2022} & RGB & 85.7 & - & - & - & - \\ 
        ~& TrajREC-Ftr. \cite{Stergiou2024} & \textit{WACV' 2024} & Skeleton & - & \underline{89.4}  & \underline{77.9}  & 68.0  & 68.2  \\ 
\midrule
        \multirow{3}*{Hybrid}&MPED-RNN \cite{Morais2019} & \textit{CVPR' 2019} & Skeleton & - & 86.3 & 75.4 & 60.6 & 61.2 \\ 
        ~& sRNN \cite{Luo2021} & \textit{T-PAMI' 2021}  & RGB & 83.5  & - & - & - & - \\ 
        ~& MoCoDAD \cite{Flaborea2023b} & \textit{ICCV' 2023} & Skeleton & - & 89.0  & 77.6  & \underline{68.3}  & \underline{68.4}  \\ 
\midrule
        \multirow{5}*{Others}& GEPC \cite{Markovitz2020} & \textit{CVPR 2020} & Skeleton & - & 58.1 & 74.8 & 53.4 & 55.2 \\ 
        ~& COSKAD-Hype. \cite{Flaborea2023a} & \textit{PR' 2024} & Skeleton & - & 87.3 & 75.6 & 64.9 & 65.5  \\ 
        ~& COSKAD-Eucli. \cite{Flaborea2023a} & \textit{PR' 2024} & Skeleton & - & 87.8  & 77.1  & 65.0  & 63.4 \\ 
        ~& EVAL \cite{Singh2023} & \textit{CVPR' 2023}  & RGB & 86.0 & - & - & - & - \\ 
        ~& OVVAD \cite{wu2024open} & \textit{CVPR' 2024 }& RGB & \underline{86.5}   & - & - & 62.9 & - \\
\midrule
        \multirow{4}*{Rec.}& MPED-RNN-Rec. \cite{Morais2019} & \textit{CVPR' 2019} & Skeleton & - & - & 74.4 & - & - \\ 
        ~& TrajREC-Prs. \cite{Stergiou2024} & \textit{WACV' 2024} & Skeleton & - & 86.3 & 73.5 & - & - \\ 
        ~& TrajREC-Pst.\cite{Stergiou2024} & \textit{WACV' 2024} & Skeleton & - & 87.6 & 75.7 & - & - \\ 
        ~& ST-PAG \cite{Rai2024} & \textit{CVPR' 2024 }& RGB & \underline{86.5} & - & - & 58.0 & - \\
        \midrule
        Rec. & FG-Diff (Ours) & - & Skeleton & \textbf{88.0}  & \textbf{90.7}  & \textbf{78.6}  & \textbf{68.9}  & \textbf{69.0}  \\ 
\bottomrule
\end{tabular}
\label{SoTA}
\end{table*}

\subsection{Experimental Setup}
Here, we introduce datasets, evaluation metric, and implementation details in brief.

\subsubsection{Datasets \& Evaluation Metric}
We evaluated our approach on five video anomaly detection benchmarks: {Avenue, HR-Avenue, HR-STC, UBnormal, and HR-UBnormal}. Avenue~\cite{Lu2013} has 16 training and 21 test clips from CUHK campus (over 30,000 frames), with normal pedestrian activities in training and anomalies like running in test clips; HR-Avenue~\cite{Morais2019} excludes non-human anomalies. HR-STC~\cite{Morais2019}, from ShanghaiTech Campus, includes 330 training and 107 test videos (over 270,000 frames, 130 anomalies), focusing on pedestrian activities and anomalies like running, excluding non-human anomalies. UBnormal, a synthetic dataset via Cinema4D, has 236,902 frames (116,087 training, 28,175 validation, 92,640 test) with 660 anomalies, designed as an open-set dataset with no anomaly overlap across splits; HR-UBnormal~\cite{Flaborea2023b} filters training anomalies and emphasizes human-related events.

 For all datasets, we adhere to the experimental protocol of prior work \cite{Flaborea2023b} and employ {Alpha Pose} \cite{Hasan2016} for initial pose estimation. This standardization results in input motion sequences characterized by a channel dimension ($C$) of $1$ (for 2D coordinates) and a fixed joint count ($J$) of $17$. The DCT embedding dimension is set to $128$, and the model is trained for $100$ epochs. For inference, we fix the number of generation (reconstruction) to ${1}$ per motion sequence. A key implementation point concerns hyperparameter selection: while {UBnormal} provides a dedicated validation set, {STC} and {Avenue} do not. Given the unified skeletal architecture, we adopt a cross-validation strategy where the test set of one serves as the validation set for the other (e.g., test set of STC is used to select the optimal hyperparameter $\lambda$ for Avenue).

Following prior work~\cite{Hasan2016}, we adopt the Area Under the Curve (AUC) as the evaluation metric. The higher AUC values indicate superior anomaly detection performance.

\subsubsection{Implementation Details} Following prior work~\cite{Flaborea2023b, Morais2019}, the data is preprocessed via segmentation and normalization. The model, comprising a perturbation generator and noise predictor with Graph Convolutional Networks (GCNs) as the backbone, is trained using the Adam optimizer with an exponential learning rate scheduler (base rate 0.01, decay factor 0.99). Anomaly scores are aggregated and smoothed using post-processing techniques~\cite{Flaborea2023b, Morais2019}. The hyperparameter \(\lambda_{p}\) is set to 0.1 for all datasets, while \(\lambda_\text{dct}\) is 0.9 for UBnormal and HR-UBnormal, and 0.1 for others.

The diffusion model employs a full {U-Net} built from ST-GCNN layers. The Encoder progressively downscales the input motion along the {joint dimension} (e.g., $17 \to 12 \to 10$) using dedicated CNN pooling layers. The subsequent Decoder reverses this process, utilizing {skip connections} from the encoder's intermediate feature maps to preserve fine-grained details during reconstruction. The diffusion time step $t$ is incorporated into every ST-GCNN layer via {Positional Encoding} with an {embedding dimension} of $128$. For regularization, a {dropout rate of $0.4$} is applied throughout the network, which follows a channel progression defined by its downsampling and upsampling paths, ultimately concluding with $C_{out}=2$ to match the 2D input coordinates.

\subsection{Comparison with State-of-the-Art Methods}

The performance comparison between the proposed method and state-of-the-art (SoTA) approaches is summarized in Table~\ref{SoTA}, where ``Pred." and ``Rec." denote {predict-based} and {reconstruction-based} methods, respectively. As discussed in Sec.~\ref{sec:intro}, existing VAD methods typically identify anomalies either by reconstructing the given motion sequence or by predicting future frames based on masked past frames. Methods that combine these two strategies are categorized as {hybrid-based methods}. We analyze the competitive results across three distinct categories: {reconstruction-based methods}, {skeleton-based methods}, and {supervised methods}.

\begin{table}[tb] \small
\centering
\caption{Comparison with supervised and weakly supervised methods. ``W.S.", ``U.S.", and ``S." denote weakly supervised, unsupervised and supervised methods, respectively. The FPS rates are measured on a single 1080Ti GPU with 11GB of VRAM. }
{
\resizebox{0.5\textwidth}{!}{ 
\begin{tabular}{lcccccc}
\toprule
\textbf{Method}  & \textbf{Training} & \textbf{Params} & \textbf{UBnormal} & \textbf{Memory} & \textbf{FPS} \\
\midrule
Sultani et al. \cite{Sultani2018} & S.  & -  & 50.3  & $<11$GB &  30.7 \\
AED-SSMTL \cite{Georgescu2021} & S. & $>$80M  & 61.3  & $<11$GB & 23.9\\
TimeSformer \cite{Bertasius2021} & S. & 121M & 68.5 & $<11$GB & 10.2 \\
AED-SSMTL \cite{Georgescu2021} & W.S. & $>$80M & 59.3 & $<11$GB & 23.9 \\
\midrule
FG-Diff (Ours) & U.S. & \textbf{556K} & \textbf{68.9}    & {$<6$GB} & \textbf{50.6} \\
\bottomrule
\end{tabular}
}}
\label{SoTA_WS}
\end{table}

\subsubsection{Reconstruction-Based Methods}
The proposed method consistently outperforms all reconstruction-based approaches across the evaluated datasets. On HR-Avenue, it achieves an AUC of 90.7, surpassing TrajREC-Pst.~\cite{Stergiou2024} by a margin of 3.1. Similarly, on HR-STC, it reaches 78.6, exceeding TrajREC-Pst. by 2.9. This improvement is attributed to the incorporation of perturbed examples during training, which enhances model robustness and mitigates overfitting, a common issue in reconstruction-based methods. By improving the distinction between previously unseen normal and abnormal samples, the proposed approach ensures more reliable anomaly detection. Additionally, by prioritizing low-frequency motion components, it alleviates the challenge of reconstructing high-frequency details, thereby leading to enhanced performance.

\begin{table}[b] \small
\centering
\caption{Ablation studies of each component in the proposed method.}
\setlength{\tabcolsep}{1.5mm}
\begin{tabular}{llll}
\toprule
\textbf{Method} & \textbf{HR-Avenue} &\textbf{HR-STC} &\textbf{HR-UBnormal} \\
\midrule
Baseline & 87.5 ($\downarrow$ 3.2) & 75.2 ($\downarrow$ 3.4) & 64.4 ($\downarrow$ 4.6) \\
Ours w/o IP & 90.4 ($\downarrow$ 0.3) & 77.4 ($\downarrow$ 1.2) & 68.7 ($\downarrow$ 0.3) \\
Ours w/ double IP & 90.7 (-) & 78.5 ($\downarrow$ 0.1) & 68.6 ($\downarrow$ 0.4) \\
Ours w/o DCT-Mask & 89.9 ($\downarrow$ 0.8) & 78.0 ($\downarrow$ 0.6) & 68.1 ($\downarrow$ 0.9) \\
\midrule
Ours & \textbf{90.7} & \textbf{78.6} & \textbf{69.0} \\
\bottomrule
\end{tabular}
\label{ablation}
\end{table}

\subsubsection{Skeleton-Based Methods}
Among skeleton-based approaches, the proposed method achieves the highest performance across all datasets, reporting AUC scores of 88.0 on Avenue, 90.7 on HR-Avenue, 78.6 on HR-STC, 68.9 on UBnormal, and 69.0 on HR-UBnormal. Compared to the previous state-of-the-art, TrajREC-Ftr.\cite{Stergiou2024}, the proposed method demonstrates consistent improvements, achieving gains of 1.3 on HR-Avenue, 0.7 on HR-STC, 0.9 on UBnormal, and 0.8 on HR-UBnormal. Reconstruction-based skeleton methods, such as TrajREC-Pst.\cite{Stergiou2024}, exhibit lower performance, as observed in the HR-STC dataset where they achieve an AUC of 75.7. In contrast, prediction-based methods, such as TrajREC-Ftr., and hybrid approaches, such as MoCoDAD~\cite{Flaborea2023b}, achieve higher scores of 77.9 and 77.6, respectively. This discrepancy arises from the limited ability of reconstruction-based methods to effectively capture temporal dynamics. The proposed approach addresses this limitation by reconstructing low-frequency components while incorporating high-frequency guidance, thereby enhancing the modeling of motion patterns. 

On the UBnormal and HR-UBnormal datasets, our method achieves AUC scores of 68.9 and 69.0, respectively, surpassing state-of-the-art methods like MoCoDAD~\cite{Flaborea2023b} (68.3 and 68.4) and TrajREC-Ftr.~\cite{Stergiou2024} (68.0 and 68.2). This improvement stems from perturbation training, which enhances the model’s representational capacity and robustness. 

\subsubsection{Comparison with Supervised and Weakly Supervised Methods}
Table \ref{SoTA_WS} evaluates the proposed method with supervised and weakly supervised methods. The proposed method outperforms existing methods with fewer parameters, demonstrating the advantage of skeleton-based methods. Even without supervision or visual information, our approach performs competitively with methods that utilize different types of supervision. Additionally, our approach boasts a significantly smaller parameter count compared to its competitors.

\begin{table}[t] \small
\centering
\caption{Robust analysis of perturbations training. ``w/ PT'' refers to the method utilizing our perturbation training, while ``w/o PT" indicates the approach without perturbation training, but with Gaussian noise augmentation instead. The symbol “$\lambda_{PI}$” denotes the perturbation intensity during inference.}
\setlength{\tabcolsep}{1mm}
\begin{tabular}{lllllll}
\toprule
\multirow{2}{*}{\textbf{$\lambda_{PI}$}} & \multicolumn{3}{c}{\textbf{w/ PT}} & \multicolumn{3}{c}{\textbf{w/o PT}} \\
\cmidrule(lr){2-4} \cmidrule(lr){5-7}
& {HR-A.} & {HR-S.} & {HR-U.} & {HR-A.} & {HR-S.} & {HR-U.} \\
\midrule
0.00  & {90.7} & {78.7}  & {68.5}  & 88.7 & 76.2& 67.4\\
0.02  & {90.6} & {78.4}  & {68.8}  & 86.9& 75.0& 66.8\\
0.05  & {90.7} & {78.2}  & {68.7}  & 84.7& 73.8& 64.2\\
0.10  & {90.7} & {78.6}  & {69.0}  & 80.3& 71.2& 63.5\\
0.15  & {90.5} & {77.9}  & {68.1}  & 76.4& 69.5& 61.3\\
\midrule
\textbf{Avg.}  & \textbf{90.6}$_{\mathbf{+7.2\%}}$& \textbf{78.4}$_{\mathbf{+5.3\%}}$  & \textbf{68.6}$_{\mathbf{+4.0\%}}$  & 83.4& 73.1& 64.6\\
\bottomrule
\end{tabular}
\label{tab:robust}
\end{table}

\subsection{Ablation Studies}
We conducted ablation studies to evaluate the impact of each component in the proposed method, comparing four models: ``Baseline" (MoCoDAD-E2E~\cite{Flaborea2023b}), ``Ours w/o IP", ``Ours w/ double IP", and ``Ours w/o DCT-Mask". The Baseline, a variant of MoCoDAD, encodes conditioned code using a trainable encoder without relying on reconstruction, thus avoiding additional hyperparameters for balancing reconstruction and prediction weights. The proposed model further leverages DCT to obtain the conditioned code. The other models are defined as follows: (1) ``Ours w/o IP" omits perturbation training while retaining other settings; (2) ``Ours w/ double IP" applies perturbation training with an input perturbation magnitude of \(\lambda\), doubling this magnitude during inference; (3) ``Ours w/o DCT-Mask" replaces the DCT-Mask with a temporal-mask, which completes motion using masked temporal dimensions for fair comparison, as the proposed model reconstructs entire motions from partial ones.

\subsubsection{Effect of DCT-Mask}
By examining the fourth row and the fifth row in Table \ref{ablation}, the results depict that the DCT-Mask is beneficial for anomaly detection, yielding improvements of 0.89\%, 0.77\%, and 1.17\%. For generative models, it is challenging to accurately reconstruct motion details. Thanks to the DCT-Mask, the proposed method can focus on generating low-frequency information with guidance of high-frequency information, leading to satisfying results.

\subsubsection{Effect of Perturbation Training}
In Table \ref{ablation}, the second row reports the results of the model without perturbation training, indicating its effectiveness for obtaining a robust model. To further verify this, we increased the magnitude of input perturbations only during testing, and the results are presented in the third row. The results remained unchanged on the HR-Avenue and declined by only 0.1\% and 0.3\% on the others, demonstrating the robustness of the model.

\begin{figure}[!t]
\centering
    \vspace{-3mm}
\subfigure[]{\includegraphics[width=0.24\textwidth]{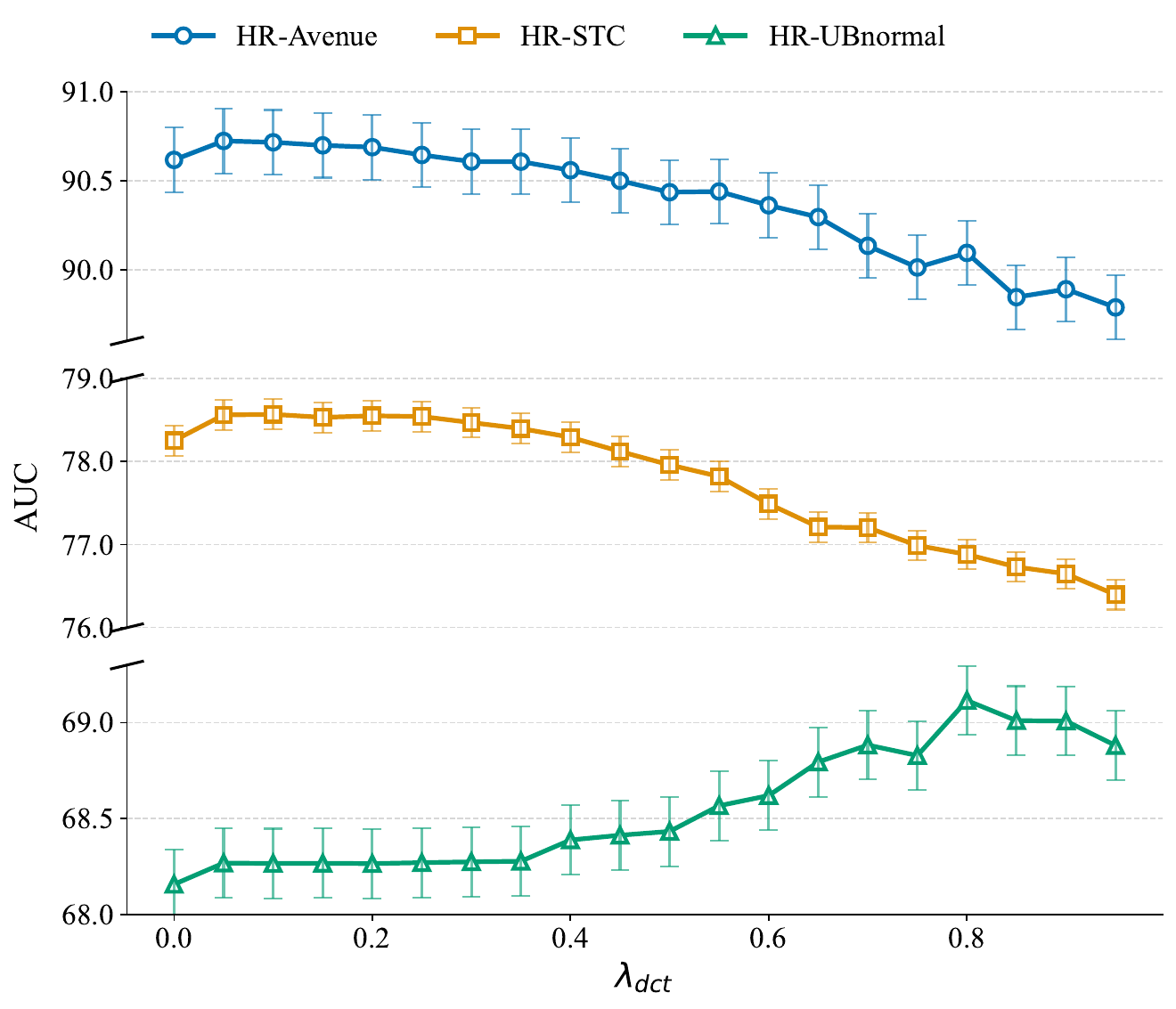}%
}
\subfigure[]{\includegraphics[width=0.24\textwidth]{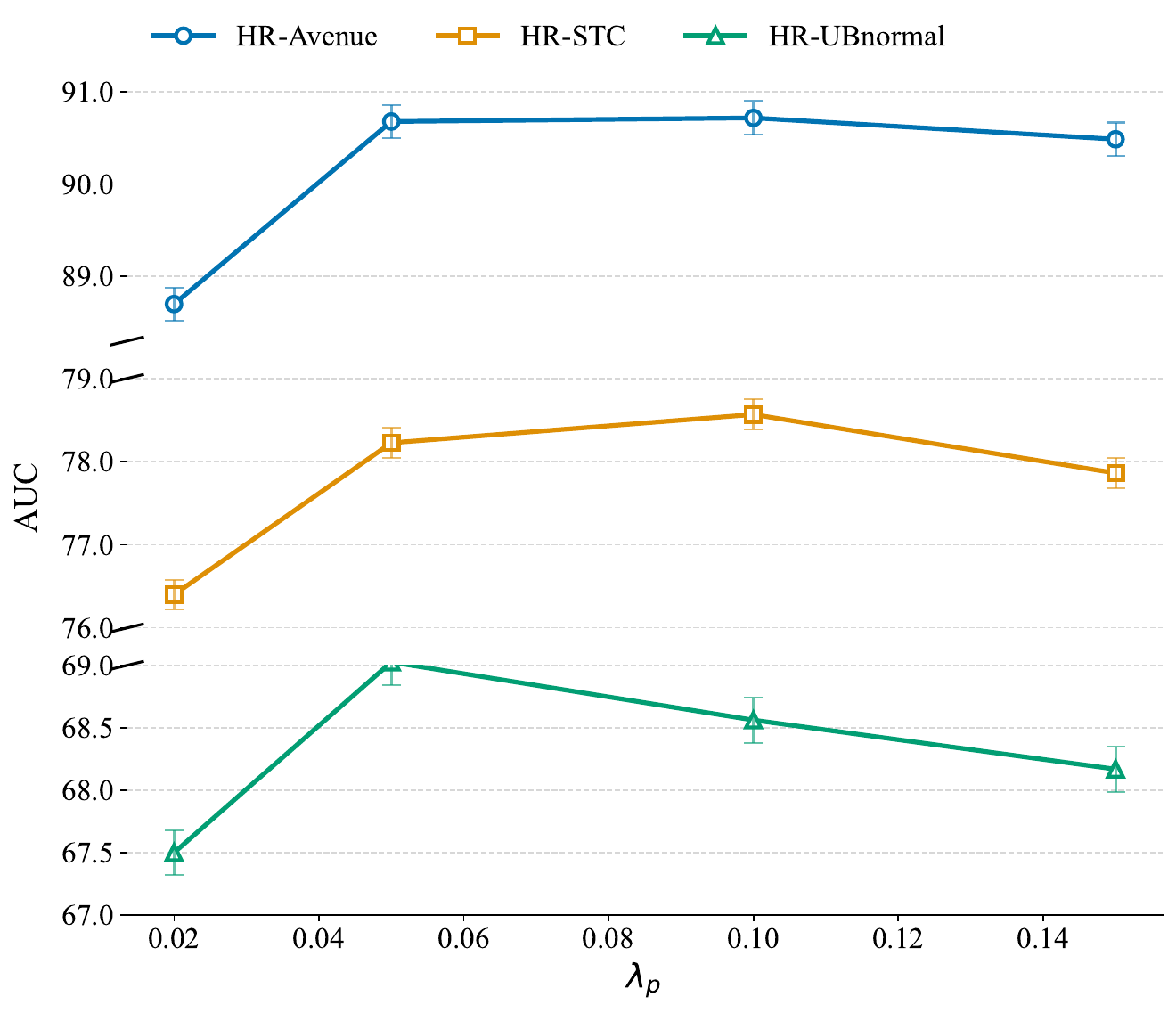}%
}
    \vspace{-5mm}
\caption{Sensitivity analyses of (a) DCT-Mask threshold $\lambda_\text{dct}$, and (b) perturbation training weight $\lambda_\text{p}$. As the model is probability-based, we performed 10 independent evaluations and reported the resulting average value.}
\label{lambda}
    \vspace{-5mm}
\end{figure}

\begin{figure}[!b]
\centering
    \vspace{-5mm}
\subfigure[]{\includegraphics[width=0.24\textwidth]{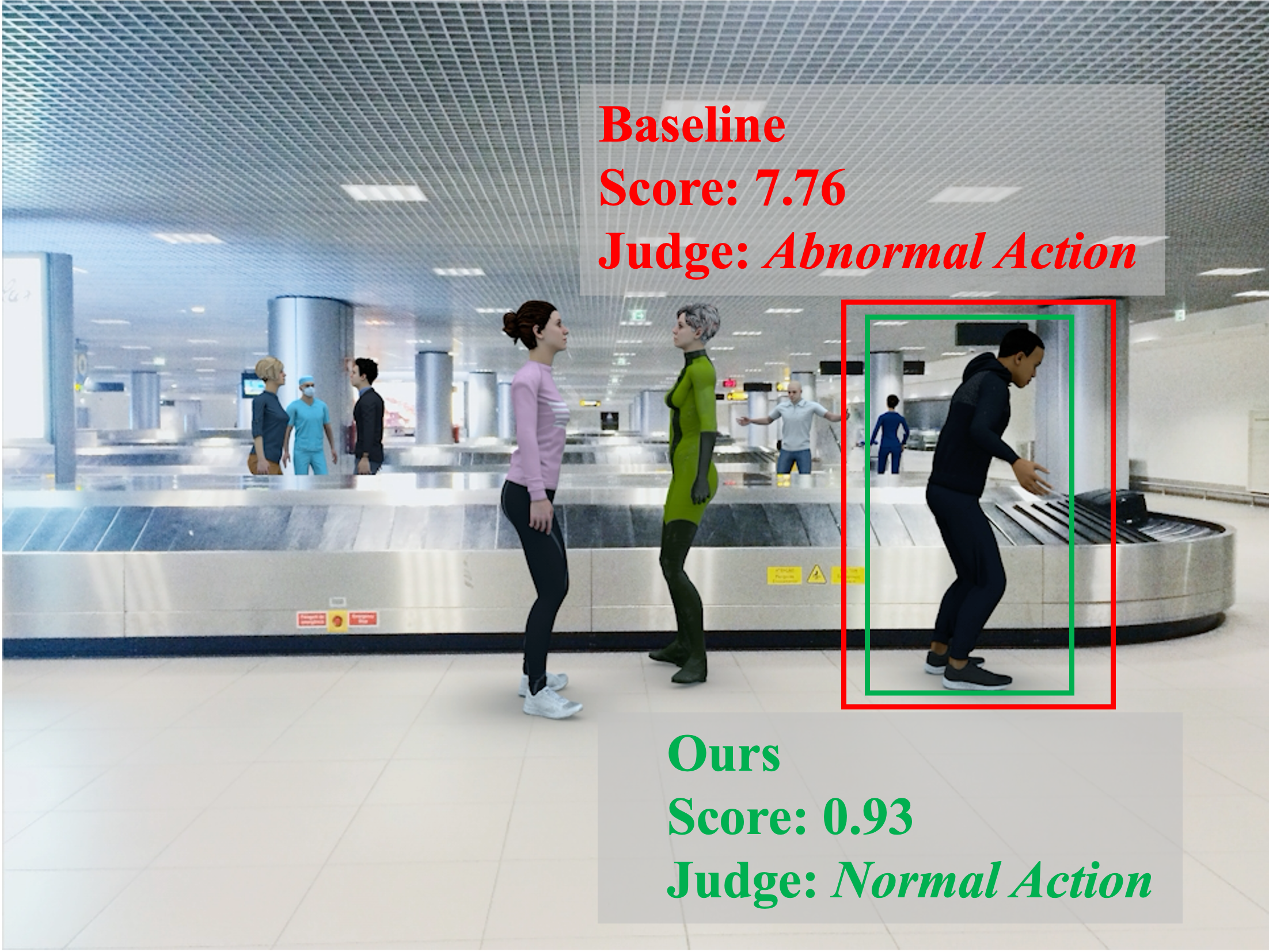}%
}
\subfigure[]{\includegraphics[width=0.24\textwidth]{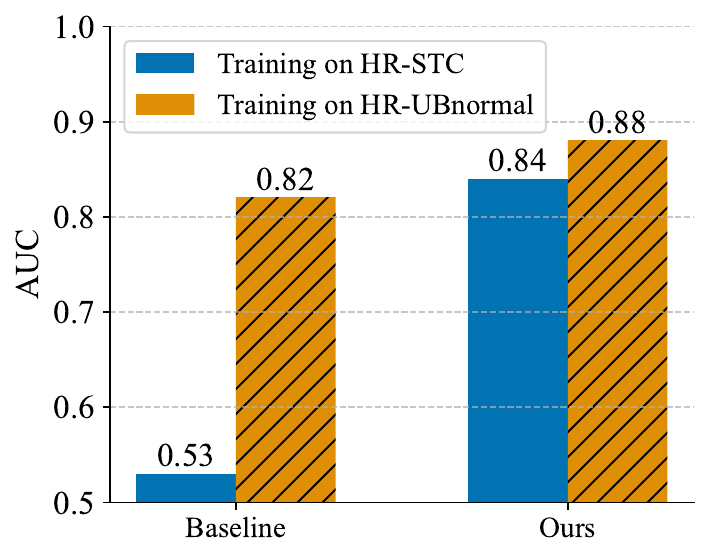}%
}
\vspace{-10pt}
\caption{Performance comparison on unseen motions from UBnormal dataset. (a) qualitative visualization; (b) quantitative results. }
\label{fig:unseen}
\end{figure}

\subsection{Analysis of Robustness and Parameters}
\subsubsection{Robustness of Perturbation Training}
In Table~\ref{tab:robust}, the ``PT" denotes whether Perturbation Training with intensity of 0.1 was utilized, $\lambda_{PI}$ represents the perturbation intensity during the inference (instead of training) for robustness analysis. For fair comparison, we use the Gaussian noise $\mathcal{N}(0,1)$ with intensity of 0.1 as data augmentation for ``w/o PT". We evaluate the robustness of perturbation training by varying the perturbation intensity for Gaussian noise $\lambda_{PI}$ during inference and comparing models trained with and without perturbation training across three HR datasets: HR-Avenue, HR-STC, and HR-UBnormal, as shown in Table~\ref{tab:robust}. The model incorporating perturbation training consistently outperforms its counterpart across all values of $\lambda_{PI}$, achieving average AUC scores of 90.6, 78.4, and 68.6, respectively, compared to 83.4, 73.1, and 64.6 without perturbation training. This corresponds to relative improvements of 7.2\%, 5.3\%, and 4.0\%. Moreover, the model trained with perturbation training maintains stable performance as $\lambda_{PI}$ increases, with AUC scores ranging from 90.7 to 90.5 on HR-Avenue. In contrast, the model without perturbation training experiences a substantial decline, with AUC scores dropping from 88.7 to 76.4 on HR-Avenue. These findings demonstrate the effectiveness of perturbation training in enhancing model robustness against varying levels of perturbation. 

To further evaluate the performance in handling unseen normal motions, we train the model on the HR-STC dataset and test it on 5083 frames of unseen normal motions and 178 frames of unseen abnormal motions sampled from the HR-UBnormal dataset. We present the qualitative  and quantitative results in Fig. \ref{fig:unseen}(a) and (b), respectively.

\begin{table}[tb] \small
\centering
\caption{The performance comparison of the violence action detection methods on the RWF-2000 dataset.}
\setlength{\tabcolsep}{1mm}
\resizebox{0.5\textwidth}{!}{
\begin{tabular}{cccccc}
\toprule
 PointNet++~\cite{PointNet2017} & DGCNN~\cite{DGCNN2019} & SPIL~\cite{SPIT2025} & ST-GCN~\cite{ST-GCN} & PDSF~\cite{Fumiaki2023} & Ours \\
\midrule
 78.2 & 80.6 & 89.3 & 83.3 & 82.5 & \textbf{91.6} \\
\bottomrule
\end{tabular}
}
\label{violence}
\vspace{-10pt}
\end{table}

\begin{figure}[b]
\centering
\includegraphics[width=0.45\textwidth]{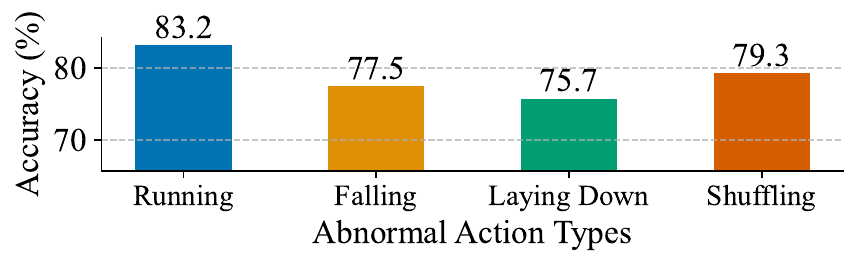}
\vspace{-10pt}
\caption{Performance analysis of specific behaviors.}
\label{fig:specific}
\end{figure}

\subsubsection{Parameter Analysis}
We analyze the impact of the DCT-Mask threshold $\lambda_{\text{dct}}$ and the perturbation training weight $\lambda_{\text{p}}$ on performance, with results detailed in Fig.~\ref{lambda}. For the DCT-Mask parameter $\lambda_{\text{dct}}$, smaller values indicate a greater reliance on the low-frequency information of the generated motion and the high-frequency information of the observed motion. As depicted in Fig.~\ref{lambda}(a), the AUC for the HR-Avenue dataset peaks at $90.7$ with $\lambda_{\text{dct}} = 0.10$, suggesting that an excessive amount of high-frequency information in the generated motions is detrimental to performance. Conversely, for the HR-UBnormal dataset, the peak AUC is observed at $\lambda_{\text{dct}} = 0.80$. This optimal value is likely attributed to the larger open-set degree of the UBnormal dataset, which is specifically designed for open-set VAD. This larger degree necessitates greater generation diversity (i.e., a higher requirement for high-frequency components in generated motions) for anomaly detection. Regarding the perturbation training weight $\lambda_{\text{p}}$, we observe that performance is insensitive to variations when $\lambda_{\text{p}} > 0$. This stability arises because perturbation training enhances the model's robustness, whereas setting $\lambda_{\text{p}}=0$ corresponds to disabling this regularization.

\subsection{Analysis of Specific Behaviors and Real-World Scenarios}

The RWF-2000 dataset \cite{RWF-2000} is a violence action detection benchmark drawn from surveillance scenarios, representing a key real-world application. For our experiment, we categorize non-violent actions as normal and violent actions as abnormal. We compare various specialized violence detection methods, with the results summarized in Table $\ref{violence}$.

To further assess specific detection capabilities, we utilized the UBnormal dataset and focused on four representative abnormal actions: \textit{running, falling, laying down}, and \textit{shuffling}. Results are shown in Fig. \ref{fig:specific}.

\begin{figure}[!t]
\centering
    \vspace{-5mm}
\subfigure[]{\includegraphics[width=0.24\textwidth]{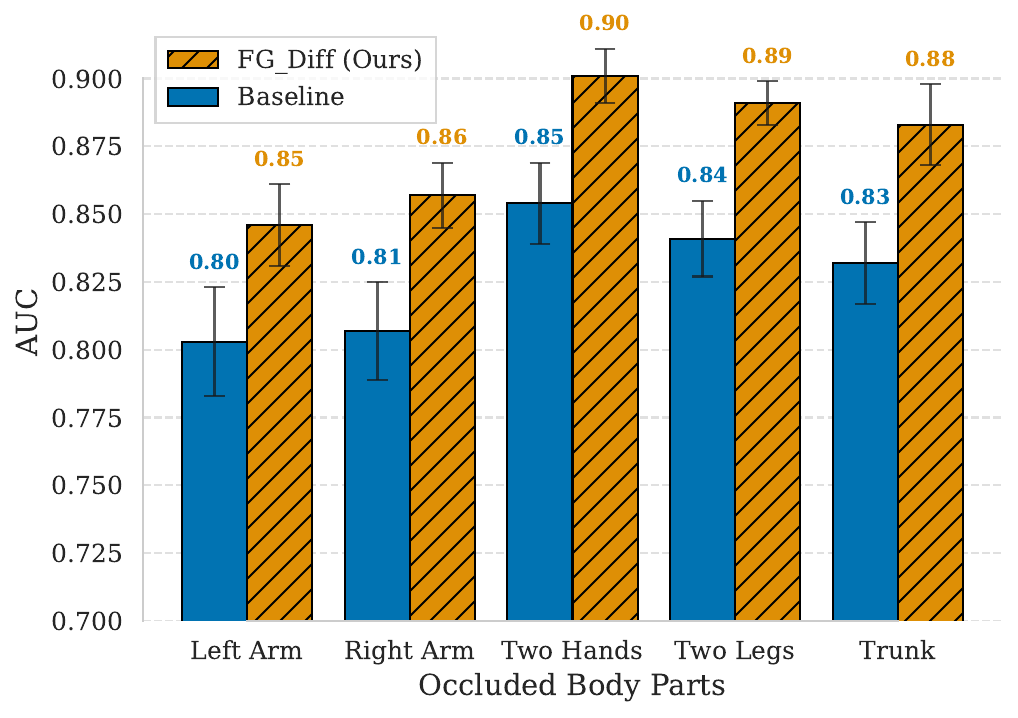}%
}
\subfigure[]{\includegraphics[width=0.24\textwidth]{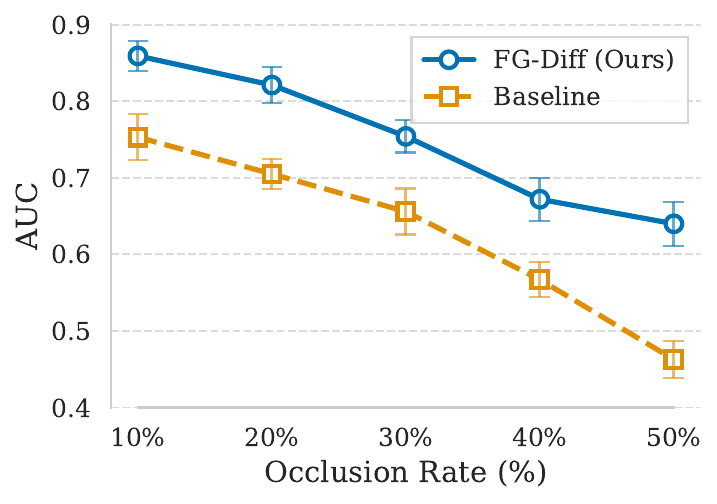}%
}
    \vspace{-5mm}
\caption{Performance of occluded actions with (a) spatial masks and (b) temporal masks. }
    \vspace{-4mm}
\label{fig:occlusion}
\end{figure}

\begin{figure}[b]
\centering
    \includegraphics[width=0.48\textwidth]{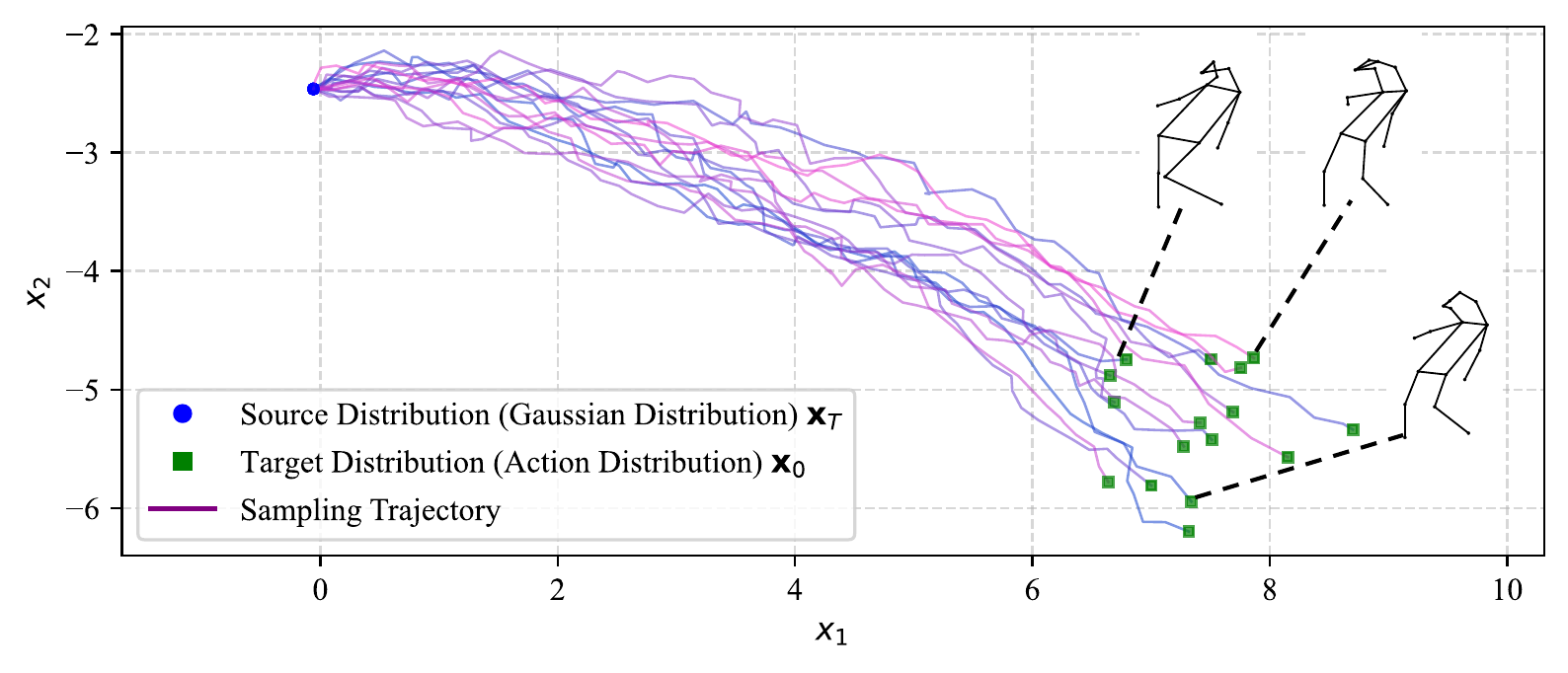}
    \vspace{-5mm} 
        \caption{Visualization of denoising process in 2D t-SNE space. The inherent stochasticity enables the reconstruction model to handle diverse actions, but the diversity of the generative poses challenges for detail reconstruction.}
    \vspace{-5pt}
    \label{fig:sde}
\end{figure}

\subsection{Robustness of Pose Occlusion}
To analyse the issue of pose occlusion, we further investigate  the performance of our model specifically in occlusion scenarios. Specifically, we adopt the experimental setup from \cite{song2019icip, Chen2023mm} to generate temporal and spatial masks that randomly occlude human action nodes and frames. For temporal occlusion, we randomly mask $k\%$ of the frames within a given motion sequence, where $k$ is varied from $10\%$ to $50\%$ with step of $10\%$. For spatial occlusion, we randomly mask specific body parts in every frame of the given motion sequence, including the left arm, right arm, two hands, two legs, and the trunk. The results on HR-Avenue are illustrated in Fig. \ref{fig:occlusion}.

We attribute the observed advantage to a combination of two factors: data inherent occlusion and the robustness of our approach. We conduct training on classical video datasets (HR-STC, HR-Avenue) where pose estimation methods like Alpha Pose \cite{fang2017rmpe} struggle with accurate estimation due to low video resolution and inherent clutter. Training the model on data already containing pose occlusion enhances its robustness. Additionally, our proposed perturbation training paradigm is also instrumental in further boosting its robustness.

\begin{figure}[!t]
\centering
\subfigure[]{\includegraphics[width=0.24\textwidth]{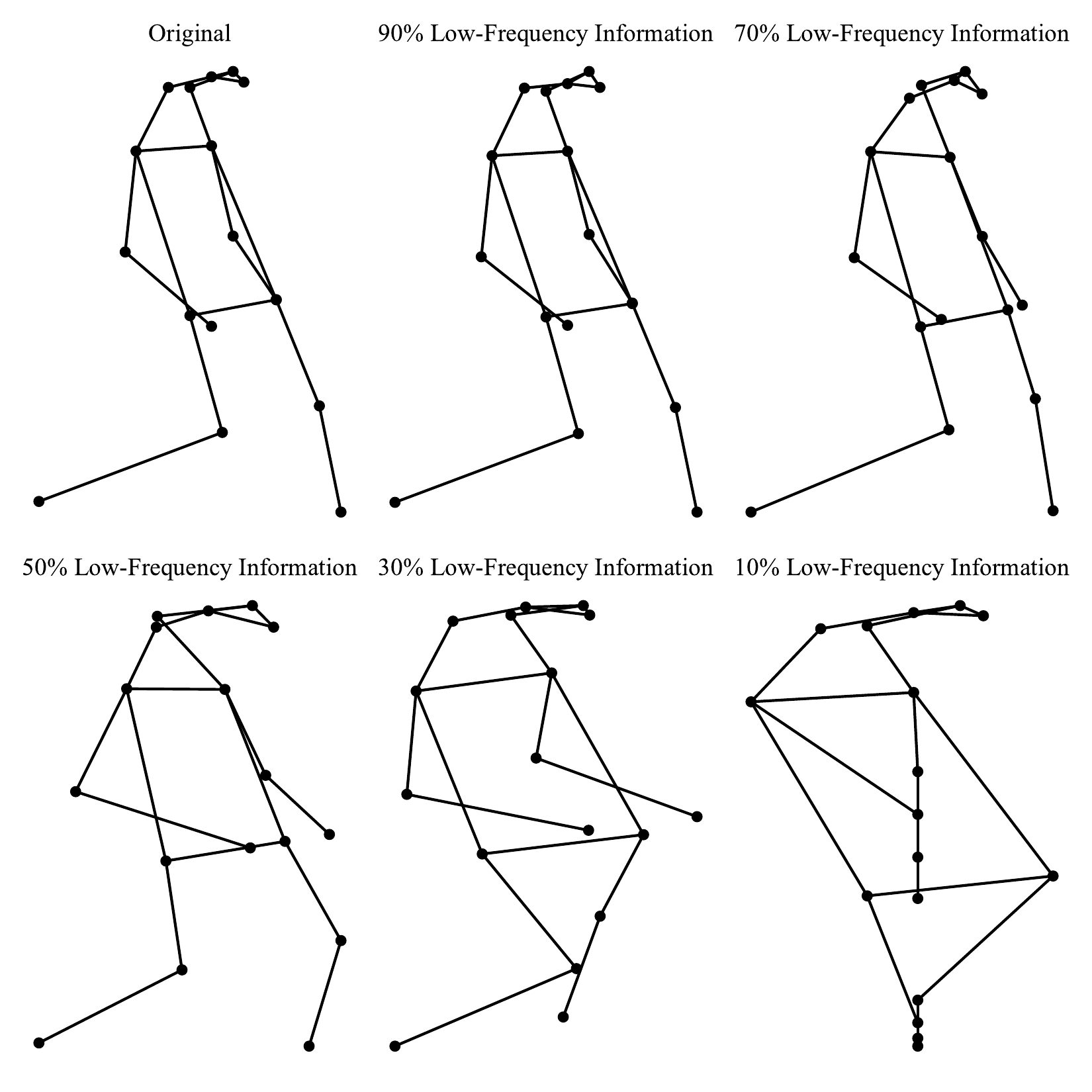}%
\label{fig:freq-s}}
\hfil
\subfigure[]{\includegraphics[width=0.24\textwidth]{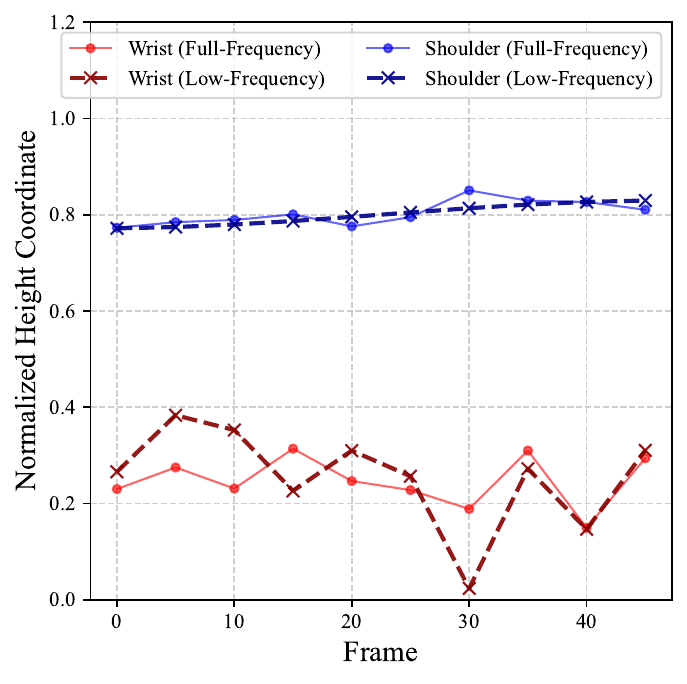}%
\label{fig:freq-t}}
\vspace{-8pt}
\caption{The low-frequency information w.r.t (a) spatiality  and (b) temporality.}
\label{fig:freq}
\end{figure}

\begin{figure}[!b]
\centering
\subfigure[]{\includegraphics[width=0.48\textwidth]{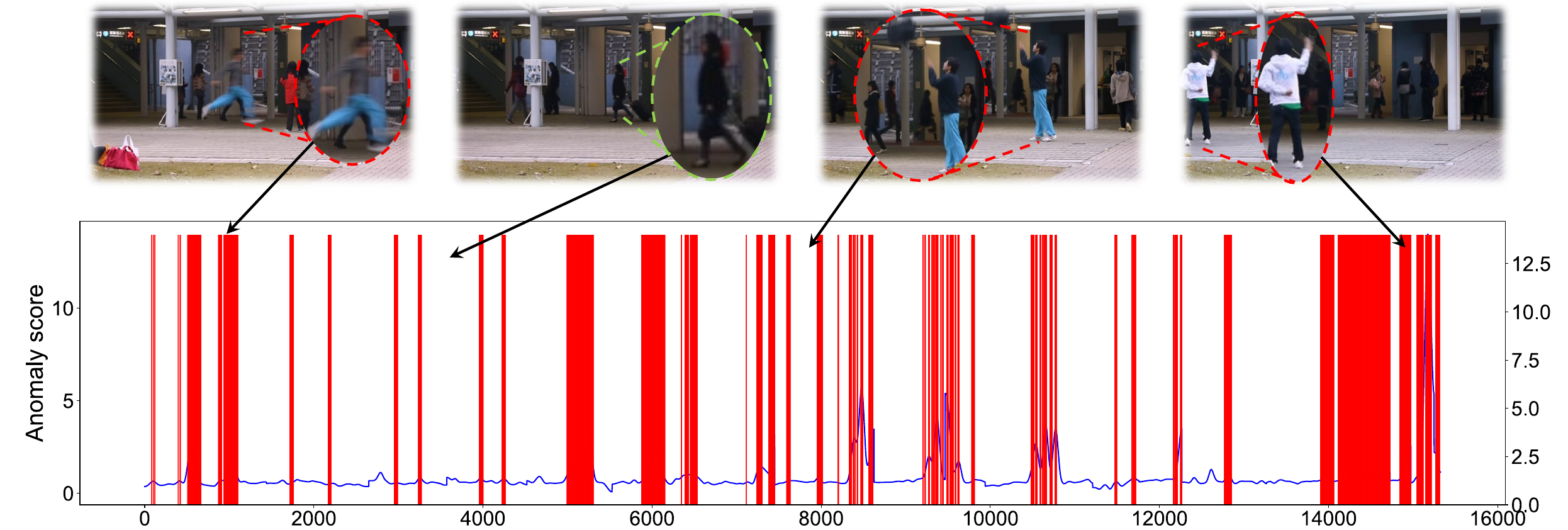}%
\label{vis:ave}}
\hfil
\subfigure[]{\includegraphics[width=0.48\textwidth]{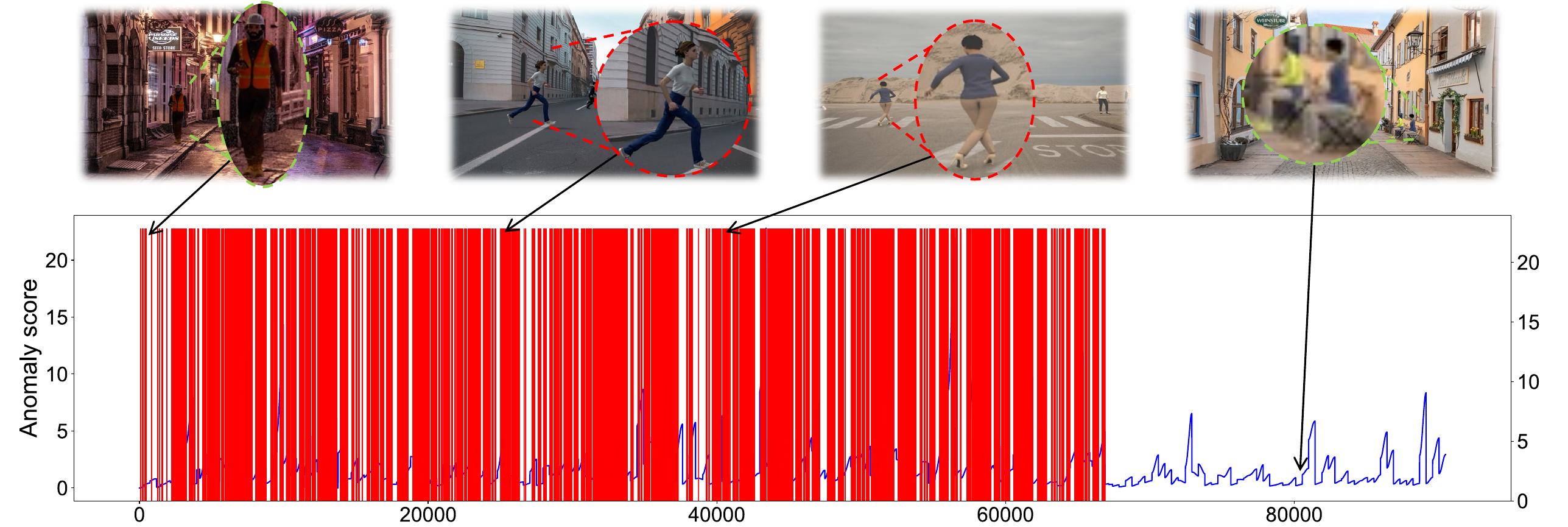}%
\label{vis:ub}}
\caption{Anomaly score curves on the Avenue and HR-UBnormal datasets. (a) Avenue dataset; (b) HR-UBnormal dataset. The horizontal axis represents the frame index, the red circles in the clip of each figure denote the abnormal events, and the green circles represent the normal ones.}
\label{vis}
\end{figure}

\subsection{Visualizations}

\noindent  \textbf{Denoising Trajectory and Frequency Analysis.} In Fig.~\ref{fig:sde}, we visualize the denoising trajectory, which demonstrates the inherent stochasticity and diversity characteristic of probability-based models, such as diffusion models. While this diversity is beneficial for handling varied outcomes, this randomness simultaneously poses a challenge for the fine reconstruction of details (high-frequency information). We further visualize the specific temporal and spatial interpretations of low-frequency information, as shown in Fig.~\ref{fig:freq}. Intuitively, low-frequency information corresponds to minimally changing components, often representing the main structure or global action, whereas high-frequency information captures rapid changes and detailed dynamics. In the {temporal dimension}, low-frequency information is predominantly concentrated in body parts with stable or minimal movement, such as the shoulders. As illustrated in Fig.~\ref{fig:freq}(b), removing the high-frequency components results in little change to the shoulder position, while the wrist position exhibits a substantial alteration. This suggests that retaining only the low-frequency information is sufficient to preserve the global motion structure. In the {spatial dimension}, low-frequency information represents the key joint nodes that primarily characterize the action itself. As demonstrated in Fig.~\ref{fig:freq}(a), maintaining only a small subset of low-frequency information is adequate to preserve the overall walking posture.

\noindent \textbf{Application Scenarios.} Fig. \ref{vis} illustrates the anomaly scores about video clips from two datasets. The results show that the proposed method can identify abnormal behaviors in the video, such as chasing, playing, and throwing.  The results show that the proposed method is sensitive to anomalies and can effectively detect anomalous events. For example, as shown in Fig. \ref{vis} (a), the anomaly score rises sharply when a man throws bags, and then, the anomaly curve returns to normal. Similarly, Fig.~\ref{vis} (b) demonstrates a peak in the anomaly score during a chasing event, highlighting the ability of the method to capture dynamic behavioral anomalies across diverse scenarios.

\section{Conclusion}
In this paper, we propose a novel frequency-guided diffusion model with perturbation training for video anomaly detection. To improve model robustness, we introduce a perturbation training strategy to expand the reconstruction domain of the model. Additionally, we use generated perturbed samples during inference to enhance the distinction between normal and abnormal motions. To tackle motion detail generation, we explore a frequency-guided motion denoising approach that leverages 2D DCT to separate high and low-frequency motion components, prioritizing the reconstruction of low-frequency components for more accurate anomaly detection. Extensive empirical results show that our method outperforms other state-of-the-art approaches.

\section{Appendix} 
\subsection{Proof of Theorem IV.1} \label{App} We provide the proof of Theorem \ref{the:1} here.

\begin{proof}
\noindent \textbf{(1)} We first prove that the perturbed motion $\hat{\mathbf{x}}^o$ is similar to observed normal motions, i.e., $\Vert \mathbf{x}^o - \hat{\mathbf{x}}^o\Vert \leq \lambda$.

Given a perturbed motion $\hat{\mathbf{x}}^o$ defined by  Eq. (\ref{eq:4})  and Eq. (\ref{eq:x}), denoted as:
\begin{equation}
    \begin{aligned}
        \delta_\phi &= \lambda_p \mathrm{sign}\big(\mathcal{G}_\phi (\mathbf{x}^o)\big),\\
        \hat{\mathbf{x}}^o &= \mathbf{x}^o + \delta_\phi.
    \end{aligned}
\end{equation}
Assuming that the dimension of motion $\hat{\mathbf{x}}^o$ is $d$, we have:
\begin{equation}
    \begin{aligned}
        \Vert \hat{\mathbf{x}}^o - \mathbf{x}^o \Vert &= \Vert \mathbf{x}^o + \delta_\phi - \mathbf{x}^o\Vert \\
        &=\Vert \lambda_p \mathrm{sign}\big(\mathcal{G}_\phi (\mathbf{x}^o)\big)\Vert \\
        &\overset{\tikz[baseline=(char.base)]{\node[shape=circle,draw,inner sep=0.5pt,font=\scriptsize] (char) {1};}}{\leq} \sqrt[d]{d} \lambda_p
    \end{aligned}
\end{equation}
where $\tikz[baseline=(char.base)]{\node[shape=circle,draw,inner sep=0.5pt,font=\scriptsize] (char) {1};}$ satisfies since the following relationship holds:
\begin{equation}
    \begin{aligned}
        \Vert \mathrm{sign}\big(\mathcal{G}_\phi (\mathbf{x}^o)\big) \Vert
        \leq\sqrt[d]{d}.
    \end{aligned}
\end{equation}
Therefore, let $\lambda=\sqrt[d]{d} \lambda_p$, the similarity relationship $\Vert \mathbf{x}^o - \hat{\mathbf{x}}^o\Vert \leq \lambda$ holds.

\noindent \textbf{(2)} Then, we prove that the perturbed motion $\hat{\mathbf{x}}^o$ will lead to an increased reconstruction error, i.e., $ \mathcal{S}(\mathbf{x}^o) - \mathcal{S}(\hat{\mathbf{x}}^o) \leq 0 $.

Eq. (\ref{eq:rec}) demonstrates that the anomaly score $\mathcal{S}({\mathbf{x}})$ is directly measured by reconstruction error, i.e., $\mathcal{S}({\mathbf{x}}) = \mathcal{L}(\mathbf{x}, \theta)$. The perturbed motion $\hat{\mathbf{x}}^o$ is obtained by:
\begin{equation}
    \hat{\mathbf{x}}^o = \mathbf{x}^o + \lambda_p \mathrm{sign}\big(\mathcal{G}_\phi (\mathbf{x}^o)\big),
\end{equation}
and $\mathcal{G}_\phi$ is optimized by:
\begin{equation}
            \max_{\phi} \mathcal{L}\Big(  \mathbf{x}^o + \lambda_p \mathrm{sign}\big(\mathcal{G}_\phi (\mathbf{x}^o)\big), \theta\Big),
\end{equation}
The following relationship holds : 
\begin{equation}
\small
\begin{aligned}
    \hat{\mathbf{x}}^o &=\arg \max_{\hat{\mathbf{x}}^o} \mathcal{L}( \hat{\mathbf{x}}^o, \theta), &\text{s.t. } \hat{\mathbf{x}}^o = \mathbf{x}^o + \lambda_p \mathrm{sign}\big(\mathcal{G}_\phi (\mathbf{x}^o)\big),
    \\\iff \hat{\mathbf{x}}^o&=\arg \max_{\hat{\mathbf{x}}^o} \mathcal{S}( \hat{\mathbf{x}}^o), &\text{s.t. } \hat{\mathbf{x}}^o = \mathbf{x}^o + \lambda_p \mathrm{sign}\big(\mathcal{G}_\phi (\mathbf{x}^o)\big).
\end{aligned}
\end{equation}
Thus, we have:
\begin{equation}
    \mathcal{S}(\hat{\mathbf{x}}^o) \geq \mathcal{S}({\mathbf{x}}^o) \iff  \mathcal{S}({\mathbf{x}}^o) - \mathcal{S}(\hat{\mathbf{x}}^o) \leq 0.
\end{equation}

The proof is completed.
\end{proof}
{
    \bibliographystyle{IEEEtran}
    \bibliography{main}
}

\vfill

\end{document}